\documentclass{article}

\usepackage{arxiv}

\usepackage{cite}
\usepackage{amsmath,amssymb,amsfonts}
\usepackage{algorithmic}
\usepackage[pdftex]{graphicx}
\graphicspath{{figures/}}
\usepackage{enumitem}

\usepackage{bm}
\usepackage{dsfont}
\usepackage{subcaption}
\usepackage{url}
\usepackage{booktabs}
\usepackage{multirow}
\usepackage{comment}
\usepackage{verbatim}
\usepackage{textcomp}
\def\BibTeX{{\rm B\kern-.05em{\sc i\kern-.025em b}\kern-.08em
    T\kern-.1667em\lower.7ex\hbox{E}\kern-.125emX}}

\title{RENT - Repeated Elastic Net Technique for Feature Selection}

\author{
 Anna Jenul \\
  Faculty of Science and Technology\\
  Norwegian University of Life Sciences\\
  \texttt{anna.jenul@nmbu.no} \\
   \And
 Stefan Schrunner \\
  Faculty of Science and Technology\\
  Norwegian University of Life Sciences\\
  \texttt{stefan.schrunner@nmbu.no} \\
  \And
 Kristian H. Liland\\
  Faculty of Science and Technology\\
  Norwegian University of Life Sciences\\
  \texttt{kristian.liland@nmbu.no} \\
  \And
 Ulf G. Indahl\\
  Faculty of Science and Technology\\
  Norwegian University of Life Sciences\\
  \texttt{ulf.indahl@nmbu.no} \\
  \And
 Cecilia M. Futsæther\\
  Faculty of Science and Technology\\
  Norwegian University of Life Sciences\\
  \texttt{cecilia.futsaether@nmbu.no} \\
  \And
 Oliver Tomic\\
  Faculty of Science and Technology\\
  Norwegian University of Life Sciences\\
  \texttt{oliver.tomic@nmbu.no} \\
}

\begin{document}
\maketitle

\begin{abstract}
Feature selection is an essential step in data science pipelines to reduce the complexity associated with large datasets. While much research on this topic focuses on optimizing predictive performance, few studies investigate stability in the context of the feature selection process. In this study, we present the Repeated Elastic Net Technique (RENT) for Feature Selection. RENT uses an ensemble of generalized linear models with elastic net regularization, each trained on distinct subsets of the training data. The feature selection is based on three criteria evaluating the weight distributions of features across all elementary models. This fact leads to the selection of features with high stability that improve the robustness of the final model. Furthermore, unlike established feature selectors, RENT provides valuable information for model interpretation concerning the identification of objects in the data that are difficult to predict during training. In our experiments, we benchmark RENT against six established feature selectors on eight multivariate datasets for binary classification and regression. In the experimental comparison, RENT shows a well-balanced trade-off between predictive performance and stability. Finally, we underline the additional interpretational value of RENT with an exploratory post-hoc analysis of a healthcare dataset.
\end{abstract}


This article is published in IEEE Access under CC BY licence, \url{www.doi.org/10.1109/ACCESS.2021.3126429}.

\section{Introduction}
\label{sec:introduction}

A predictive task involves a dataset consisting of $N$-dimensional row vectors $X=(\bm{x}_1^T,\dots,\bm{x}_{I}^T)\in\mathbb{R}^{I\times N}$ and an associated vector of target values $\bm{y} = (y_1,\dots,y_{I})\in \mathbb{T}^I$, where the target space $\mathbb{T}$ may represent a set of classes (classification task) or a subset of the real numbers (regression task). In this study, our focus lies on generalized linear models (GLMs), which model the target as a linear combination of the inputs with weights $\bm{\beta}\in\mathbb{R}^N$, followed by a transformation. The columns of the data matrix describe object characteristics, denoted as features. Since data acquisition techniques evolve steadily, situations where the number of features $N$ exceeds the number of objects $I$ often occur. In such setups, mathematical obstacles, like spurious correlations and multicollinearity issues causing model overfitting, trigger the necessity to reduce the number of features by using some feature selection approach\cite{liu2010}. These issues are characteristic of various domains, including healthcare \cite{Ashour,Li2020}, biomedicine \cite{saeys2008}, text mining \cite{bai2018particle} and botany \cite{Moghimi}. A successful feature selection approach will decrease the model complexity, improve the model stability and provide more useful model interpretations.  

A feature selector $\theta_F$ decomposes the data space into a direct sum of selected features ($V_1$) and non-selected features ($V_2$) according to the given feature set $F\subset\{1,\dots,N\}$,
	$$\mathbb{R}^N=V_1 \oplus V_2,~\text{s.t.} V_1 \cong \mathbb{R}^{\vert F\vert} ~\text{and}~ V_2 \cong \mathbb{R}^{N-\vert F\vert},$$ 
and projects all objects from $\mathbb{R}^N$ to the subspace $V_1$, i.e.
	$$ \theta_F: \mathbb{R}^{N}\rightarrow V_1,~\theta_F(\bm{x})=\text{proj}_{V_1}(\bm{x}). $$
The goal of good feature selection is to determine the feature set $F^\star$, which enables a predictive model to obtain the most accurate prediction. Predictive quality is measured using a metric $q\left(\hat{\bm{y}},\bm{y}\right)$, such as F1 score, where $\hat{\bm{y}},\bm{y}\in \mathbb{T}^{I_\text{test}}$ denote the vectors containing predicted target values $\hat{\bm{y}}$ after feature selection and ground truth target values $\bm{y}$, both referring to a set of test data $X_\text{test}$ of size $ \lvert X_\text{test}\rvert = I_\text{test}$. An optimal feature set $F^\star$ is characterized by
$$ F^\star = \underset{F\subset\{1,\dots,N\}}{\text{arg max}} q(\hat{\bm{y}}^F,\bm{y}). $$

A taxonomy of feature selection techniques distinguishes between filter, wrapper, and embedded approaches. Filter approaches rank features by an importance criterion, such as mutual information or correlation coefficients between features and target variables. Baseline filters include the Fisher score \cite{bishop:fisherscore} and the Laplacian score \cite{He05} as well as algorithms from the relief family \cite{kira1992}. Approaches like mRMR \cite{peng05} or the stratified feature weight method \cite{Chen18} aim to resolve the issue that correlated and redundant features are not well handled by classical filters \cite{cherrington2019}. A combination of different filter approaches is suggested in \cite{Haq}.
Wrapper approaches select features concerning their prediction performance. By training supervised models on different subsets of the entire feature set, the subset delivering the most accurate predictions on a test set is chosen. This strategy often causes overfitting issues and high computational costs \cite{loughrey2004}. Prominent wrapper approaches are forward/backward selection, such as recursive feature selection \cite{guyon2002gene}, and heuristic searches like simulated annealing or genetic algorithms \cite{Brownlee11}.

The third category of feature selection methods, embedded feature selection, integrates the selection step directly into the learning algorithm. A class of embedded methods, which is particularly important in this work, comprises regularization for GLMs: During parameter estimation, regularization terms are added as penalties to the target function. While the well-established LASSO \cite{hastie09} uses an L1 term $\lambda_1(\bm{\beta})=\vert \bm{\beta}\vert$ for this purpose and delivers a sparse parameter vector, L2-regularization $\lambda_2(\bm{\beta})=\Vert \bm{\beta} \Vert_2$ handles multicollinearities by pulling the L2-norm of the parameter vector $\bm{\beta}$ towards zero. The effects of both regularization terms are combined in the elastic net $\lambda_{enet}(\bm{\beta})$\cite{Hastie15}, defined as
\begin{equation}
\lambda_{enet}(\bm{\beta}) = \gamma [\alpha\lambda_1(\bm{\beta})+ (1-\alpha)\lambda_2(\bm{\beta})],
\label{eq:enetpars}
\end{equation}
with parameters $\alpha\in [0,1]$ and $\gamma$ to weight the regularization terms and to define the regularization strength, respectively.
Other representatives of embedded feature selection models are tree-based models, such as decision trees or regression trees. Ensembles of tree-based architectures are called random forests \cite{tuv2009}. Graph-based approaches together with elastic net regularization further play a key role in recent works \cite{cui19,cui2021,cui2021fused}, where the authors demonstrate a graph-based structurally interacting elastic net method incorporating pairwise relationships between objects via a feature graph, or \cite{pang19}, proposing a solution for $\ell_{2,0}$-norm regularized feature selection via linear discriminant analysis.

Most feature selection approaches suffer from the phenomenon that minor changes in the random initialization or train-test-split of the model lead to major variations in the selected feature set---this issue is referred to as lack of stability and is investigated in \cite{nogueira2017} and \cite{bommert2017}. In agreement with \cite{bovelstad2007} and \cite{xu2011}, the authors argue that L1 regularisation on GLMs is generally unstable. They claim that the issue can be resolved by investigating ensemble feature selection, where $\theta_F$ is derived from a set of independently trained (elementary) feature selectors $\theta_{F_1},\dots,\theta_{F_K}$, such that $\theta_F = \phi\left(\theta_{F_1},\dots,\theta_{F_K}\right)$. The operator $\phi$ acts as a meta-model based on information from the elementary models $\theta_{F_k}$, $k=1,\dots,K$. A basic approach is to build such a meta-model by counting the frequency of selection for each feature across all feature sets $F_k$, expressed by $$F^{\star} = \left\{i \in \{1,\dots,N\} : \tau_1(i) = \frac{1}{K}\left\vert \{k: i\in F_k \} \right\vert \geq t_1\right\},$$ where $t_1\in [0,1]$ is a scalar representing a minimum selection frequency threshold. This approach assumes that each elementary feature set $F_k$ consists of a subset of important features (with correspondingly higher probabilities for being selected), and a small, random subset of unimportant features. The final, selected feature set $F^\star$ is less likely to contain unimportant features than each of the elementary feature sets $F_k$. Hence, model stability is increased as shown by Meinshausen and Bühlmann \cite{meinshausen2010}, who propose such a feature selector named stability selection. Even though the stability selection framework is intuitive and reasonable, the corresponding feature weights may be small---not significantly different from zero---or have alternating signs across the elementary models. Thus, features might be selected although resulting in ambiguous or contradictory information and hence, deteriorating interpretability and predictive performance. This means that further insights into the predictive power of features have to be gained from the distribution of weights, which is not considered by Meinshausen and Bühlmann \cite{meinshausen2010}.

The present work suggests the novel repeated elastic net technique (RENT) for feature selection. RENT is based on the idea of model ensembles discussed in \cite{meinshausen2010}. Besides merely calculating the frequency of each feature, we also focus on the empirical distribution of the feature weights resulting from elastic net regularized models. Thereby, we extend the model ensemble framework to combine three rigid selection criteria: 1) how often is a feature selected?; 2) to which degree do the feature weights alternate between positive and negative values?; 3) are feature weights significantly different from 0? The final feature selection of RENT consists of the features that satisfy all three selection criteria. When required, the RENT framework can be extended with additional custom criteria to refine the feature selection process according to the user's a priori insights and requirements. By taking elastic net regularization into account, RENT aims at optimizing predictive performance and model stability simultaneously. In contrast, the concept of stability selection focuses on model stability as the primary target. We suggest a hyperparameter selection procedure based on the Bayesian information criterion (BIC) to balance the number of features and the predictive performance. In the experiments section, we explore and evaluate RENT extensively using real-world datasets for both classification and regression problems. In addition, we use the information provided by the ensemble models for an exploratory post-hoc analysis with statistical tools, including principal component analysis. Our implementation (in Python code) is publicly available and published in the Journal of Open Source Software \cite{RENT2021}.

\section{Repeated Elastic Net Technique for Feature Selection}\label{sec:REN}
In this section, we present the methodological concept of RENT which relies on regularized logistic regression for binary classification problems and regularized linear regression for regression problems. We introduce the idea of elastic net regularization combined with repeated training of machine learning models on unique subsets of the training data to investigate feature selection stability. Finally, we define three quality metrics that influence the feature selection.

\subsection{Ensemble Training and Selection Criteria}
\label{subsec:RENT_sel_crit}

\begin{figure*}[ht]
    \centering
        \includegraphics[width = 0.9\textwidth]{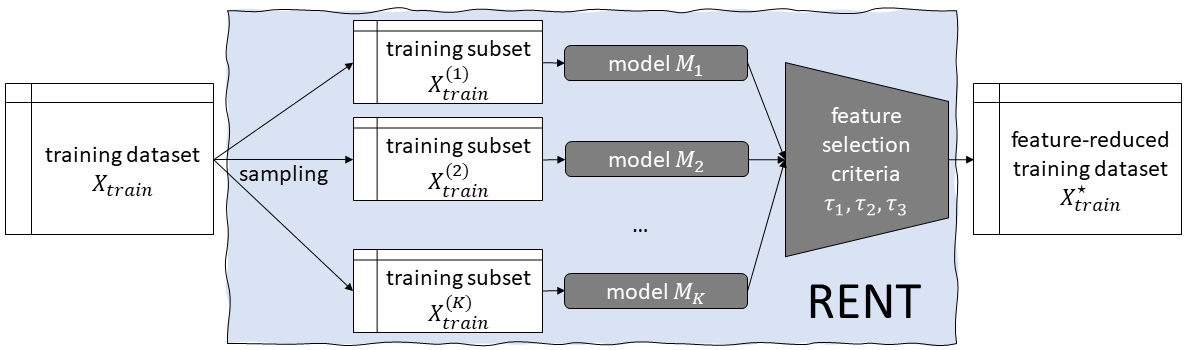}
    \caption{RENT feature selection pipeline. When using other feature selection methods, the blue frame is replaced by the other feature selectors, listed in Table \ref{tab:methods}.}
    \label{fig:overviewFigure}
\end{figure*}

Given a set of training data $X_{train}= \{\bm{x}_i: i=1,\dots,I_{train}\}$ where $\bm{x}_i$ denotes an object from the $N$-dimensional feature space,
our concept builds on sampling $K$ unique i.i.d. (independent and identically distributed) subsets $X_{train}^{(k)}\subset X_{train}$ of size $I_{train}^{(k)}$. As shown in Fig. \ref{fig:overviewFigure}, a regularized GLM $M_k$ is trained on $X_{train}^{(k)}$ for each $k=1,\dots K$. 

The evaluation of each model $M_k$ is performed on the validation set $X_{val}^{(k)} = X_{train}\setminus X_{train}^{(k)}$ (here, $\setminus$ denotes the set difference operator). To further improve robustness, we include the option to introduce more variation across the $K$ models, by randomly varying the number of objects drawn from $ X_{train}^{(k)}$ between the models within user-specified limits.
For each feature $n$ in $X_{train}$, $n=1,\dots N$, we observe the trained weights $\beta_{k,n}$ throughout models $M_k$, $k=1,\dots,K$. For the purpose of feature selection, we acquire relevant information about the importance of feature $n$ across all models from $\bm{\beta}_n = \left(\beta_{1,n},\dots,\beta_{K,n}\right)$. All such vectors $\bm{\beta}_n, n=1,\dots, N$, are aggregated in a matrix $\bm{B}$ of dimension ($K \times N$). 
Since all models comprise L1 regularization terms, the vectors of feature weights $\bm{\beta}_n$ are typically sparse. However, entries are not constant due to 1) variations in the training subsets and 2) numerical deviations in the parameter optimization. Hence, a straightforward measure of feature relevance is the relative frequency $c(\bm{\beta}_n)$, counting how often a feature was selected on average across the $K$ models or, in other words, calculating the relative frequency as an estimate of the probability for the parameter of the $n$-th feature to be non-zero:
\begin{equation}
    c(\bm{\beta}_n)= \frac{1}{K}\sum\limits_{k=1}^K \mathds{1}_{[\beta_{k,n}\neq 0]}.
\end{equation}
Furthermore, we observe two other empirical summary statistics of the feature parameter estimate distributions in the rows of $B$: the feature-specific mean $\mu(\bm{\beta}_n)$ and variance $\sigma^2(\bm{\beta}_n)$ of the feature weights
\begin{align}
    \mu(\bm{\beta}_n) &= \frac{1}{K}\sum\limits_{k=1}^K \beta_{k,n},\\
    \sigma^2(\bm{\beta}_n) &=\frac{1}{K-1}\sum\limits_{k=1}^K (\beta_{k,n}-\mu(\bm{\beta}_n))^2. 
\end{align}

In general, we consider the $n$-th feature to be a candidate for selection in RENT if
\begin{enumerate}
\item $c(\bm{\beta}_n)$ is large, i.e. the feature is selected in many of the $K$ elastic net models;
\item the estimates in $\bm{\beta}_n$ resulting from the $K$ models do not alternate much between positive and negative signs (stability);
\item the mean of distribution resulting from the $K$ parameter estimates in $\bm{\beta}_n$ is significantly non-zero.
\end{enumerate}
These three simple and transparent requirements may be formulated in corresponding mathematical expressions, to form three quality metrics for assessing a feature $n$:
\begin{align}
\tau_1(\bm{\beta}_n) &= c(\bm{\beta}_n); \\
\tau_2(\bm{\beta}_n) &= \frac{1}{K}\bigg\vert \sum\limits_{k=1}^K \text{sign}(\beta_{k,n})\bigg\vert; \\
\tau_3(\bm{\beta}_n) &= t_{K-1}\left(\frac{\vert \mu(\bm{\beta}_n)\vert}{\sqrt{\frac{\sigma^2(\bm{\beta}_n)}{K}}}\right),
\end{align}
where $t_{K-1}(.)$ denotes the cumulative distribution function of Student's $t$-distribution with $K-1$ degrees of freedom. 

Considering the second quality metric $\tau_2(\bm{\beta}_n)$, the ideal case for feature $n$ would be that all weights have the same sign---either all positive or all
negative. In case of constant signs among all weights, $\tau_2(\bm{\beta}_n)$ equals $\tau_1(\bm{\beta}_n).$ Though, for a considerably large $K$, we should expect that at least slight sign variations for some features may occur. $\tau_2(\bm{\beta}_n)$ simply allows the user to define a required minimum proportion of the parameter estimates to have the same sign.
The third quality metric $\tau_3(\bm{\beta}_n)$---identifying consistently high model parameter estimates---is chosen such that it corresponds to the well-known statistical Student's $t$-test with rejection of the null hypothesis
$$ H_0: \mu(\bm{\beta}_n) = 0.$$
In case that the null hypothesis holds, the test statistic $$ T=\frac{\mu(\bm{\beta}_n)}{\sqrt{\frac{\sigma^2(\bm{\beta}_n)}{K}}} $$ will follow a Student's $t$-distribution with $K-1$ degrees of freedom. The deployed term evaluates the probability of the test statistic under the $H_0$-distribution and thus, provides a thresholding at the chosen level of significance.

In order to define feature selection criteria from quality metrics $\tau_1(\bm{\beta}_n)$, $\tau_2(\bm{\beta}_n)$ and $\tau_3(\bm{\beta}_n)$, we introduce corresponding cutoff values $t_1, t_2, t_3\in [0,1]$. Specifically, a feature $n\in F$ is added to the selected feature set $F^\star$, if it satisfies all three criteria: $\tau_i\geq t_i, \forall i\in \{1,2,3\}$. Further criteria can be included by the user if necessary. In the provided setup, these quality metrics may be considered as hyper-parameters of the RENT method, allowing the user to regulate the feature selector, by tuning the thresholds $t_1$, $t_2$ and $t_3$. The cardinality of the selected features $F^\star$ will increase, if any of these thresholds are decreased and vice versa. All three metrics, $\tau_1$, $\tau_2$ and $\tau_3$, are bounded by the interval $[0,1]$, which facilitates the specification of appropriate thresholds. Since $\tau_3$ can be associated with a Student's $t$-test, the threshold $t_3$ for a 5\% or 1\% significance level, corresponds to the thresholds $t_3=0.95$ and $t_3=0.99$, respectively.

\subsection{Hyperparameter selection}
\label{subsec:hyperparameter_selection}
RENT involves hyperparameters at different stages of the method: before training the elementary models, regularization parameters $\gamma$ and $\alpha$ control the restrictiveness of the feature selection in the ensemble, followed by the parameters $t_1,t_2,t_3$ determining the final feature set. Thereby, the latter cutoff parameters are (a) dependent on the choice of the regularization parameters, and (b) have mutual dependencies.

Hyperparameter selection is commonly performed using an additional validation dataset or cross-validation---both options are not optimal for RENT, since a validation subset would reduce the number of objects in a high-dimensional dataset even further, and cross-validation would add a substantial computational burden to the procedure. Thus, we deploy an alternative approach from statistical model selection: the Bayesian information criterion (BIC) delivers a trade-off between the information content (quantified as the likelihood) of the model and the model complexity in terms of the number of estimated parameters \cite{BIC}. BIC is defined as 
\begin{equation}
    \text{BIC} = -2\log \hat{\mathcal{L}} + I_\text{train}\log \rho,
\end{equation}
where $\hat{\mathcal{L}}$ denotes the estimated likelihood of the predictive model, and $\rho$ denotes the number of estimated model parameters. In contrast to similar information measures like the Akaike information criterion (AIC), BIC is known for stronger penalization of model complexity leading to a lower number of selected features, which is favorable in the case of RENT. By minimizing BIC, models with high information content and low complexity are favored. In ordinary linear regression models and other standard GLMs, the number of estimated model parameters equals the number of variables, i.e., features, plus one parameter for the offset $\beta_0$; thus, we set $\rho = \vert F\vert + 1$. The likelihood $\hat{L}$ can be determined from the distribution assumptions of the GLM model, such as the normal distribution of errors in the ordinary least squares regression model, resulting in the sum of squared errors (SSE) as negative log-likelihood function.

RENT uses a two-step hyperparameter estimation procedure: a grid search for regularization parameters $\alpha$ and $\lambda$ with BIC as target function is performed on the full training dataset first (step 1). Then, the RENT ensemble is trained given the best regularization parameter combination. Finally, in step 2 another grid search for cutoff parameters $t_1,t_2,t_3$ is performed using the same concept as in step 1.

\subsection{Training runtime complexity of RENT}
\label{subsec:computational_complexity}
Since RENT is an ensemble method built on GLMs as elementary models, the runtime complexity of RENT is expressed as a multiple of the runtime complexity of GLMs, denoted by $\mathcal{O}_{GLM}$. In essence, $\mathcal{O}_{GLM}$ depends on the applied type of GLM, the parameter optimization algorithm, and the implementation. For instance, a runtime complexity of $ \mathcal{O}_{GLM} = \mathcal{O}(N^3 + I_{train}\cdot N^2)$ is reported for Lasso by reducing the computation to solving a least squares regression problem\cite{efron2004}. Variants using iterative algorithms are rather judged by the overall experimental runtime and the runtime complexity per update cycle, while the number of iterations is hard to determine a priori---such information is provided for GLMs with elastic net regularization in \cite{friedman2010}.

Given the first variant, RENT runs an ensemble comprising $K$ independent GLMs, each trained on a number of $N$ features, which delivers a complexity of $$ \mathcal{O}\left(KN^2\cdot (N+I_{train}^{(K)})\right),$$
where $I_{train}^{(K)} < I_{train}$ denotes the sample size of each subset during RENT training.
In addition, hyper-parameter tuning requires training $c$ GLMs, where $c$ is a constant given by the number of level combinations for regularization and cutoff parameters, resulting in $$\mathcal{O}\left(cN^2 \cdot (N+I_{train})\right).$$
In total, an upper bound to the full runtime complexity of RENT is given by \begin{equation}
    \mathcal{O}\left( (K+c) \cdot N^2 \cdot (N+I_{train}) \right).
\end{equation}

\section{Experiments}\label{sec:Experiments}
We demonstrate the potential of RENT as a feature selection method through experiments on multiple datasets. First, we verify the overall concept in a validation study in Section \ref{subsec:validation_study}. Second, we evaluate the performance of RENT in comparison with seven feature selection methods and a baseline elastic net regularized model, in Section \ref{subsec:comparison}.
Based on one dataset, we illustrate how the stability of RENT behaves compared to the stability of established ensemble methods based on the number of unique elementary models $K\in\mathbb{N}$. 

\subsection{Experimental Setup and Datasets}
\label{subsec:datasets}

Experiments are conducted on multivariate datasets from various domains, including real-world data and synthetic data for both binary classification (class.) and regression tasks (reg.); datasets are listed in Table \ref{tab:datasets}. 
 The size of each dataset is denoted via the number of features $(\# feat)$ and the number of objects $(\# obj)$ divided into train and test sets (train/test). Train-test-splits are performed by stratified random sampling. Further, the \textit{class balance} indicates the percentage of class representation for each classification dataset (train/test).

 The broad selection of use cases, including high-dimensional datasets, demonstrates the flexibility and applicability of RENT. Simulated datasets (c0) and (r0) were produced using scikit-learn \cite{scikit-learn} functions \textit{make\_classification} and \textit{make\_regression}, respectively. For the MNIST dataset, two binary classification problems are defined by restricting the classes: MNIST$_{cl_1, cl_2}$ indicates that only instances from classes $cl_1$ and $cl_2$, where $cl_1, cl_2\in\{0,\dots,9\}$, were used, ignoring objects from other classes.

\begin{table*}
\caption{Classification (class.) and regression (reg.) datasets used for evaluation of the feature selection methods.}
\centering
\resizebox{\textwidth}{!}{
\begin{tabular}{l c r r r r r r }
        \toprule
        \multirow{2}{*}{task} & \multirow{2}{*}{acronym} & \multirow{2}{*}{dataset} &  \multirow{2}{*}{source} & \multicolumn{2}{c}{size} &  \multicolumn{2}{c}{class balance}\\
         & & & & \# feat. & \# obj. (train/test) & \% class 0 (train/test) & \% class 1 (train/test)\\
        \midrule
        \multirow{6}{*}{class.} & c0 & synthetic dataset & simulated & 1000 & 175/75 & 0.50/0.47 & 0.50/0.53 \\
        \cmidrule(l){2-8}
        &c1 & MNIST$_{0,1}$ &  \multirow{2}{*}{\cite{lecun10}} & 784 & 12$\,$665/2115 & 0.47/0.46 & 0.53/0.54 \\
        &c2 & MNIST$_{4,9}$ &  & 784 & 11$\,$791/1991 & 0.50/0.49 & 0.50/0.51 \\
        \cmidrule(l){2-8}
        &c3 & Breast cancer Wisconsin &  \cite{wolberg1990} & 30 &  399/170 & 0.62/0.65 & 0.38/0.35\\
        \cmidrule(l){2-8}
        &c4 & Dexter text classification &  \cite{NIPS2004}& 20$\,$000 & 300/300 & 0.5/0.5 & 0.5/0.5 \\
        \cmidrule(l){2-8}
        &c5 & OVA Lung &  \cite{stiglic2010} & 10$\,$935 & 1083/462 & 0.92/0.92 &0.08/0.08\\
        \midrule
        \multirow{2}{*}{reg.} & r0 & synthetic dataset & simulated & 1000 & 175/75 & - & -\\
        \cmidrule(l){2-8}
        &r1 & Milk protein dataset & \cite{liland2009} & 6179 & 45/45 & - & -\\
        \bottomrule
\end{tabular}
}
\label{tab:datasets}
\end{table*}

A feature selector is trained on $X_{train}$, then the training data $X_{train}$ is projected into the subspace spanned by the selected features. This column-reduced training dataset is denoted by $X_{train}^{\star}$.
In our experiments we train an unregularized linear/logistic regression model $M^{\star}$ on $X_{train}^{\star}$. 
Evaluation is based on the predictive performance obtained from 
$M^{\star}$ on the previously unseen test data $X_{test}$. It is important to note, however, that it may be necessary to use regularization for the model $M^{\star}$ to avoid overfitting, especially if the reduced $X_{train}^{\star}$ has more features than objects. 

\subsection{Evaluation Metrics}
\label{subsec:evaluation_criteria}

We use two different measures for quantitative evaluation of the prediction performance in classification settings: F1 score (F1) and Matthews correlation coefficient (MCC) \cite{Matthews75}. The F1 score represents the harmonic mean of precision (PR) and recall (RC). Denoting the entries of the confusion matrix by TP (true positive), FP (false positive), FN (false negative), and TN (true negative), the performance measures are defined as follows:

\begin{align}
    PR &= \frac{TP}{TP+FP};\\
    RC &= \frac{TP}{TP+FN};\\
    F1 & = 2\cdot \frac{PR\cdot RC}{PR+RC};\\
    MCC &= \frac{TP \cdot TN - FP \cdot FN}{\sqrt{P\cdot(TP+FN)\cdot(TN+FP)\cdot N}},
\end{align}
where $P=TP+FP$ and $N=TN+FN$ are the sums of the predicted positives and negatives, respectively.
Note that F1 scores can be calculated for both class labels, depending on which class is considered as "positive". F1 is more appropriate than accuracy for imbalanced class distributions because the larger class dominates the latter. A disadvantage of F1 is that it does not take into account TN. Therefore, MCC provides more representative results if both classes are equally relevant in the prediction problem and the number of TN objects is high. F1 score, precision, and recall are bounded between $[0,1]$, where $0$ represents a complete disagreement between predicted and actual class, and $1$ denotes a perfect match. MCC is bounded between $[-1,1]$, where $-1$ denotes that all objects are classified incorrectly, $0$ indicates complete randomness, and $1$ denotes correct classification of each object, respectively.

For regression problems, we evaluate the root mean squared error of prediction (RMSEP) on the test dataset $X_{test}$ with cardinality $I_{test}$, defined as
\begin{equation}
    RMSEP = \sqrt{\frac{1}{I_{test}} \sum\limits_{i=1}^{I_{test}} (y_i - \hat{y}_i)^2},
\end{equation}
and the coefficient of determination ($R^2$) \cite{draper1998}
\begin{equation}
    R^2 = 1 - \frac{\sum\limits_{i=1}^{I_{test}} (y_i - \hat{y}_i)^2}{\sum\limits_{i=1}^{I_{test}} (y_i-\bar{y})^2},
\end{equation}
where $y_i$ represents the true output of object $\bm{x}_i$, $\hat{y}_i$ represents the prediction of $y_i$ and $\bar{y}$ represents the mean of the outputs $y_i$, $i\in \{1,\dots,I_{test}\}$. While $RMSEP$ is always non-negative we seek its minimization. $R^2$ on the other hand may take negative values but has an upper bound of $1$ (associated with perfect predictions) and we therefore seek its maximization.

Besides predictive performance, selection stability is assessed using a measure suggested in \cite{nogueira2017} evaluating the different outcomes of multiple feature selection runs in a combinatorial way.
Specifically, the suggested measure computes a ratio between the sample variance of observed feature frequencies and the theoretical variance, given that the feature selector is stable (null hypothesis). The authors clarify that their measure fulfills five consistency criteria and is asymptotically bounded by the interval $[0,1]$, where $1$ denotes optimal stability. Their concept of measuring feature selection stability by aggregating multiple independently trained models underlines the relevance of our ensemble approach and supports the idea to achieve stability by combining $K$ independent feature selection model runs.

\subsection{RENT Hyperparameter Selection}
\label{subsec:hs}
\begin{figure}[ht]
    \centering
    \includegraphics[width=0.7\columnwidth, clip]{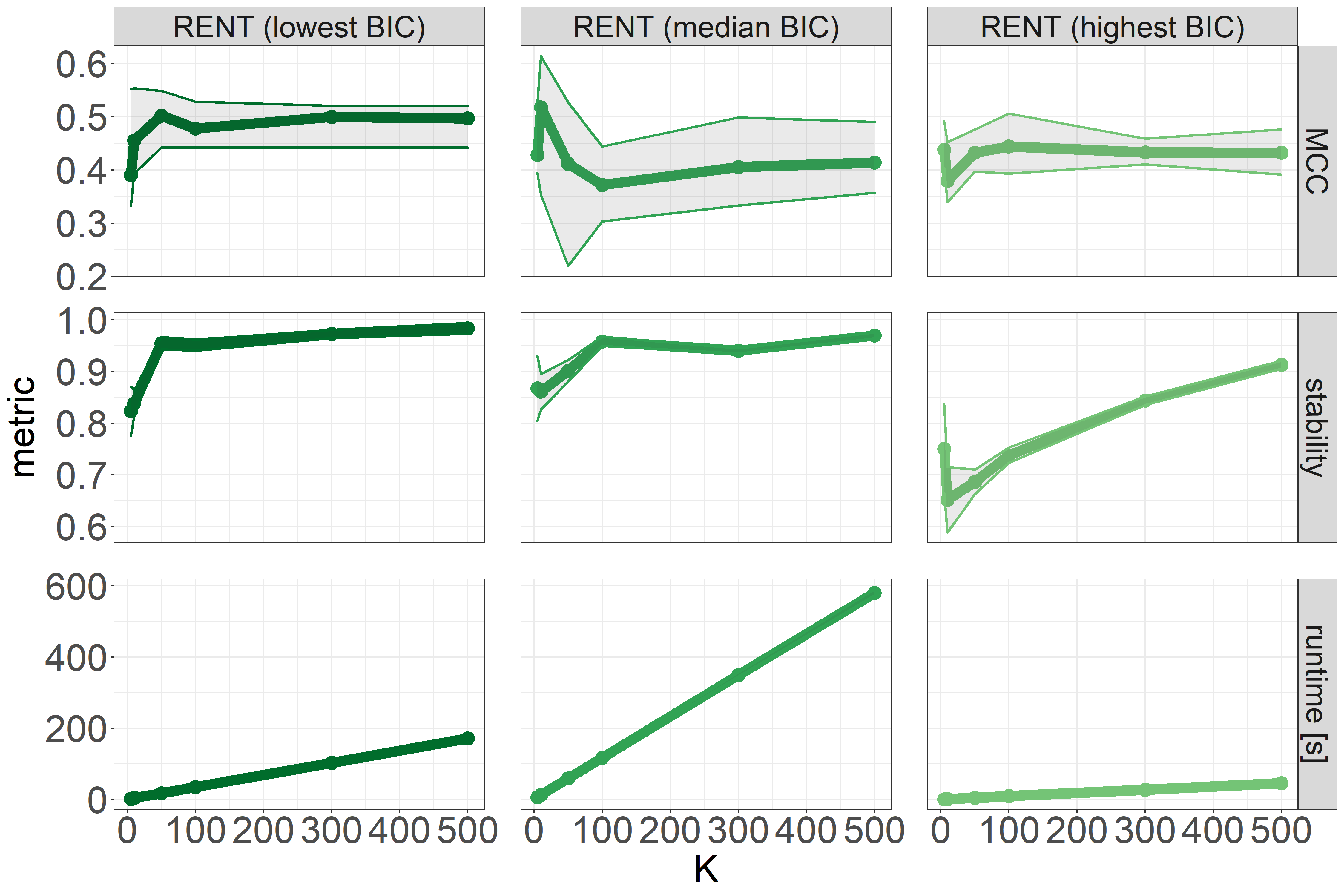}%
    \caption{Comparison of RENT with different hyperparameter setups (elastic net regularization and cutoff) for dataset c0 to varying numbers of ensemble models $K$. Each setup is evaluated in 30 independent runs. The plot shows mean values (bold line) and empirical $2.5\%$ (lower line) and $97.5\%$ (upper line) quantiles.}
    \label{fig:stability_RENT}
\end{figure}

In Section \ref{subsec:hyperparameter_selection} we introduce hyperparameter selection for both the elastic net modeling and the three cutoff parameters based on the BIC. More precisely, we evaluate the elastic net hyperparameter combinations of $\gamma\in \{1e^{-2},1e^{-1},1\}$ and $\alpha \in \{0,0.1, 0.25,0.5,0.75,0.9,1\}$. To find the best combination concerning BIC, we train a single logistic/linear regression model with each pairwise combination of hyperparameters $\gamma$ and $\alpha$ on the training dataset. 
After determining optimal elastic net parameters $\gamma$ and $\alpha$ for a given dataset in Table \ref{tab:datasets}, all ensemble models $M_1, \dots, M_K$ in RENT are trained with these parameters. Once all $K$ models are fitted on their respective training subsets $X_{train}^{(k)}$, we select the cutoff hyperparameters $t_1, t_2$ and $t_3$ with BIC, in the same way as for the elastic net hyperparameter search. For this purpose we perform a grid search on $t_1\in [0.2,1]$ with stepsize $0.05$, $t_2$ within the same range and $t_3 \in \{0.9, 0.95,0.975,0.99\}$, representing different significance levels in the $t$-test. A comparison of three different hyperparameter settings based on the lowest, median, and highest BIC values is shown for dataset c0 in Fig. \ref{fig:stability_RENT} for a varying number of elementary models $K\in\{5,10,50,100,300,500\}$. For each hyperparameter setting $(t_1, t_2, t_3)$ leading to the lowest, median and highest BIC, respectively, 30 independent runs of RENT are carried out. Each run is conducted on the same training dataset but with a distinct (random) model weight initialization.  Performance is measured via the MCC; runtimes are given in seconds and refer to one single run for each method. Across all 30 independent runs, the mean is calculated together with the empirical $2.5\%$ and $97.5\%$ quantiles (corresponding to a two-sided 5\% confidence interval) for stability, performance, and runtime, respectively.

We can observe that, as expected, the setup of RENT with optimal hyperparameters (lowest BIC) outperforms those settings with median and highest BIC and achieves the highest stability. Especially RENT based on hyperparameter settings $(t_1, t_2,t_3)$ with the highest BIC is unstable, even though the performance remains in an acceptable range. Regarding runtime, it takes about 600 seconds for RENT with median BIC to run a single model for $K=500$, which is much longer than for the other two settings. A reason for this might be a harder optimization task for specific hyperparameter combinations where it takes more steps for the logistic regression model to converge.

In summary, we observe the following behavior of RENT with lowest BIC at an increasing number of models $K$: 
\begin{itemize}
    \item on average, good MCC and stability are achieved simultaneously, even with low $K$;
    \item as expected, stability increases significantly from 0.75 for $K=5$, saturating at a value close to 1;
    \item average MCC shows little change from $K=100$ to $K=500$;
    \item runtime increases linearly with the number of trained models.
\end{itemize} 
Hence, our results support the analysis in \cite{saeys2008} that repeated use of regularized elastic net models is useful to achieve stable and reproducible results, while keeping the predictive performance at a high level. Our observations further indicate, that no major benefit can be achieved by increasing the number of models to more than approximately 100 with respect to the observed metrics on the given dataset. Therefore, $K = 100$ seems to be a valid default regarding the trade-off between stability and time for the datasets used in this study. If computation time is critical, the user may set $K$ to a lower number but needs to consider that the distribution of the weights may be insufficiently covered and that this may have an impact on the stability of feature selection.

Alternatively, instead of using BIC, the user may set hyperparameters $\gamma$ and $\alpha$ manually or use cross-validation to obtain a customized trade-off between predictive performance, stability, and the number of selected features. Note that this approach may be more subjective and that the computational cost can be higher than using BIC, especially if cross-validation is used. 

\subsection{Validation Study of Features Selected with RENT}
\label{subsec:validation_study}

\begin{table}[ht]
\centering
\caption{Prediction results per dataset of the validation study (MCC for c0-c5,  $R^2$ for r0 and r1) showing the total number of features, the number of features selected with RENT ($\Delta$), and the performance metrics. The column $RENT$ gives the MCC/$R^2$ of a predictive model trained after feature selection. \textsuperscript{a}$\geq 0.99$.}

\begin{tabular}{c | r r r r r }
        \toprule
          data- & \# features & \# features  & \multicolumn{3}{c}{MCC / $R^2$} \\
          set & total & $\Delta$ & RENT & (VS1) & (VS2)  \\
         \midrule
         c0 & 1$\,$000 & 12 & 0.50 & 0.07 & -0.02 \\
         c1 & 784 & 37 & 0.99\textsuperscript{a} & 0.98 & 0.00 \\
         c2 & 784 & 124 & 0.95 & 0.81 & 0.00 \\
         c3 & 30 & 4 & 0.96  & 0.81 & -0-01 \\
         c4 & 20$\,$000 & 5 & 0.33 & 0.00 & 0.00 \\
         c5 & 10$\,$935 & 16 & 0.91 & 0.23 & 0.00 \\
         \midrule
         r0 & 1$\,$000 & 28 & 0.99 & -0.19 & -0.90 \\
         r1 & 6$\,$179 & 8 & 0.71 & - & - \\
        \bottomrule
\end{tabular}
\label{tab:validation}
\end{table}

To demonstrate the validity of features selected with RENT, we apply two validation study setups (VS1) and (VS2). In
(VS1) we draw random features, while in (VS2) we randomly
permute labels of the test dataset. In both cases, we build
logistic regression models, predict on an unseen test dataset
and compare MCC scores to predictions based on features
selected by RENT. The comparisons are performed via one-sided Student's $t$-tests where the null hypotheses claim that
the MCC of RENT is lower or equal to the average MCCs
obtained from (VS1) or (VS2), respectively. For regression
datasets, the analog procedure is applied using $R^2$ as a
quality metric. Both tests are conducted at a
significance level of $0.05$.  
\begin{enumerate}
    \item[(VS1)]Compare a number of $\ell\in\mathbb{N}$ randomly selected feature sets, representing inefficient feature selections, to the features selected by RENT. The steps of the procedure are:
        \begin{enumerate}
        \item{sample $\ell$ independent, random feature subsets from $X_{train}$, containing $\Delta$ features each}, where $\Delta$ corresponds to the number of features selected by the RENT approach
        \item{train a new model for each of the $\ell$ feature sets by restricting $X_{train}$ to those features}
        \item{predict the labels of $X_{test}$ with each of the $\ell$ models and compute MCCs}
        \item{perform a Student's $t$-test, assuming as null hypothesis that the MCC value obtained from RENT is drawn from the same distribution}
        \end{enumerate}
    \item[(VS2)] Compare the predictive performance of a model based on features selected with RENT on the real $X_{test}$ labels, to the predictive performance of $\ell$ randomly permuted labels of $X_{test}$. The steps of the procedure are:
        \begin{enumerate}
        \item{train a model on $X_{train}$ with the features selected with RENT}
        \item{randomly permute $y_{test}$ $\ell-$times and compute the average MCC over the $\ell$ permutations}
        \item{perform a Student's $t$-test, assuming as null hypothesis that the MCC value obtained from RENT is drawn from the same distribution}
        \end{enumerate}
\end{enumerate}

\begin{figure}[ht]
    \centering
        \includegraphics[width = 0.7\columnwidth]{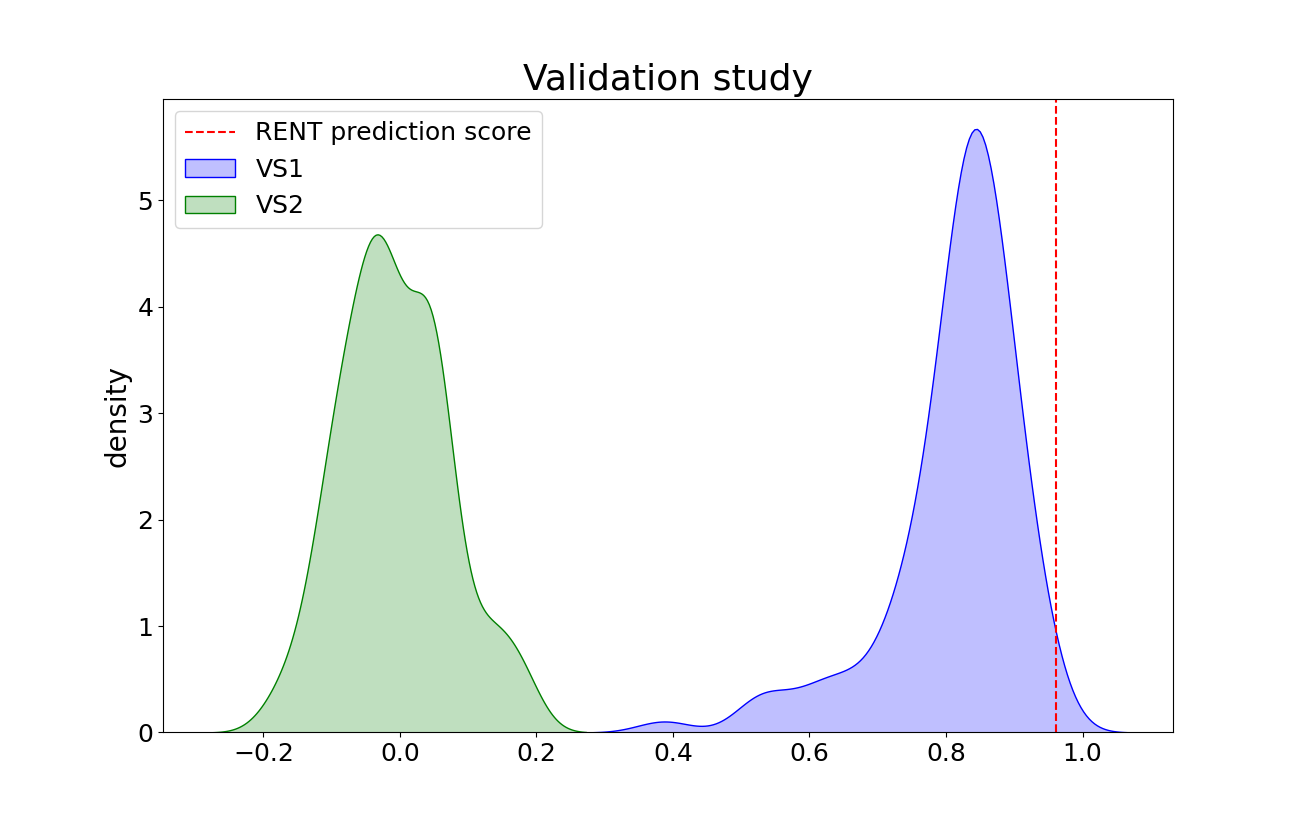}
    \caption{Empirical distributions of MCC scores in studies (VS1) and (VS2) on dataset c3 represent the validation study's results. The red line indicates the MCC based on RENT features.}
    \label{fig:Validation_Study}
\end{figure}

Performance results from (VS1) and (VS2) provide a reliable indicator of whether models based on features selected by RENT perform better than models based on randomness. Table \ref{tab:validation} shows the average MCC of (VS1) and (VS2) in the columns $MCC/ R^2$. All corresponding $p$-values from the Student's $t$-tests are significantly lower than 0.05, mostly below $1e^{-15}$, where $\ell$ equals 100. Since the standard deviation of the mean decreases with the sample size, a higher explanatory power of the Student's $t$-tests can be achieved by setting $\ell$ to a larger value. However, the runtime increases linearly in $\ell$. The estimated densities of (VS1) and (VS2) for the Breast cancer Wisconsin dataset (c3) are plotted in Fig. \ref{fig:Validation_Study}. 

In general, these two validation studies are not limited to RENT and may be applied to other feature selection methods and metrics, as well. 
Overall, the null hypotheses in both VS1 and VS2 were rejected for all datasets, indicating that RENT performs significantly better than models based on randomness as described in the validation approaches. The experimental results in Table \ref{tab:validation} show that (VS2) is close to zero across all datasets, as one would expect from the experimental setup. (VS1) performance is similar to the performance of RENT models for datasets c1. This fact may be explained by the individual information content of each feature: especially two-class subsets extracted from MNIST contain many mutually or highly correlated features. Therefore, many different feature combinations lead to good predictions.

\begin{figure}[ht]
    \centering
    \begin{subfigure}{0.49\columnwidth}
    \centering
        \includegraphics[width=0.5\columnwidth, trim = 0 0 0 0, clip]{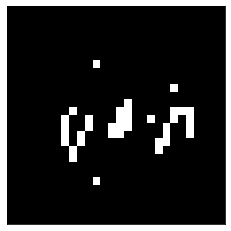}%
        \caption{Class 0 versus class 1.}
    \end{subfigure}
    \hfill
    \begin{subfigure}{0.5\columnwidth}
    \centering
        \includegraphics[width=0.49\columnwidth, trim = 0 0 0 0, clip]{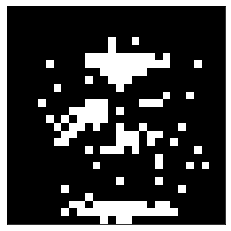}%
        \caption{Class 4 versus class 9.}
    \end{subfigure}
    \caption{Visualization of features selected by RENT for the MNIST datasets c1 (class 0 versus class 1) and c2 (class 4 versus class 9) from the 28 $\times$ 28 images of the numbers. Selected features are colored in white.}
    \label{fig:Mnist}
\end{figure}

RENT results for MNIST (datasets c1 and c2) are visualized in Fig. \ref{fig:Mnist}. We observe that 1) different features are relevant for distinguishing the class pairs 0-1 and 4-9 and 2). features relevant for 0-1 are typically located in the center of the image, whereas those relevant for 4-9 are more distributed across the image. Overall, distinguishing between 4 and 9 is more complex. Therefore, the number of selected features is much higher in this case than when classifying the numbers 0 and 1.

\subsection{Comparison of RENT with Established Feature Selectors}
\label{subsec:comparison}
The validation study in Section \ref{subsec:validation_study} showed that RENT is a valid feature selection approach for all datasets used in this study. 
Hence, we compare RENT to the methods listed in Table  \ref{tab:methods} as follows: 1) seven established feature selectors applied to classification datasets; 2) five feature selectors applied to regression datasets;  3) a baseline logistic/linear regression model $M^\circ$ with elastic net regularization \cite{hastie09}. For each feature selector, software implementations are publicly available. To compare RENT to traditional filter methods, we consider the Laplacian score (L-score) \cite{He05}, Fisher score (F-score) \cite{bishop:fisherscore}, mRMR \cite{peng05}, and a representative of the relief family, reliefF \cite{kononenko1997}. Specifically, we select the top features according to the scores provided by each filter method. Further, we study the behavior of recursive feature elimination (RFE) \cite{guyon2002gene} representing a wrapper based approach. Finally, our comparison also involves two prototypes of state-of-the-art ensemble feature selectors: stability selection (StabSel) \cite{meinshausen2010} and the random forest \cite{breiman2001random}, which can be used for both classification (RFC) and regression (RFR) problems.

\begin{table}[ht] 
\caption{Established feature selection techniques representing benchmarks for the experimental evaluation of RENT.}
\centering
\begin{tabular}{r r}
        \toprule
        method & implementation \\
        \midrule
        elastic net ($M^\circ$)   & Python \cite{scikit-learn}\\
        StabSel & R  \cite{Hofner:stabsel}\\
         RFC/RFR & Python \cite{scikit-learn}\\
        L-score & R \cite{Lapl_Impl}\\
        F-score & R \cite{Lapl_Impl}\\
        mRMR & R \cite{mrmre-impl}\\
        ReliefF & Python \cite{reliefF_randy}  \\
        RFE &  Python \cite{scikit-learn} \\
        \bottomrule
\end{tabular}
\label{tab:methods}
\end{table}
In contrast to other methods, RENT and $M^\circ$ share the advantage that the user does not have to specify the size of the selected feature set as input. Instead, the number of selected features is indirectly controlled via the elastic net regularization parameters $\gamma$ and $\alpha$. Similarly, for StabSel the exact number may be specified optionally. Otherwise, it is determined indirectly by a cutoff and an upper bound per-family error rate (PFER) \cite{Hofner:stabsel}. For fair performance comparison of the remaining investigated methods, the size of the selected feature set 
is set to the number of features returned by RENT, denoted as $\Delta$. 

For StabSel, we perform a 5-fold cross-validated grid search to estimate adequate parameter settings. The elementary feature selection method is the logistic regression model with L1 regularization; the number of models equals $K$. Furthermore, we perform a grid search for stability selection on the interval $[0.6,0.9]$ for the cutoff value and $[0.05, 0.95]$ for the PFER value, with a $0.05$ step size each. In our study, the random forest serves as a filter, delivering a ranking of the features. The features with the highest ranks are selected and used as input for $M^{\star}$ where the number of the selected features corresponds to the number of features $\Delta$ selected by RENT. To fit the random forest model, we set the number of unique trees to $K$. Other parameters are set to the defaults. 
Computations are performed on standardized train datasets. 
All model parameters used for the established methods, such as the neighborhood graph construction in L-score or the step size in RFE, are set to the default values, except for c4 and c5, where the step size is increased to 100 in order to obtain results in a moderate runtime. The regularization parameters $\gamma$ and $\alpha$ for $M^\circ$ are set to those used for RENT. The results for all datasets and methods are provided in Table \ref{tab:resultsClass} for binary classification datasets and in Table \ref{tab:resultsReg} for regression datasets.\footnote{The GitHub repository \url{https://github.com/annajenul/RENT_article_results} stores example code to reproduce the results.}

\begin{table*}[ht]
\caption{F1 scores and MCC results for classification datasets. \textsuperscript{a} $\geq0.99$, \textsuperscript{b} returned error.}
\centering

\footnotesize
\begin{tabular}{clllllllll}
 & \multicolumn{1}{c}{$M^\circ$} & \multicolumn{1}{c}{RENT} & StabSel & \multicolumn{1}{c}{RFC} & \multicolumn{1}{c}{L-score} & \multicolumn{1}{c}{F-score} & \multicolumn{1}{c}{mRMR} & \multicolumn{1}{c}{reliefF} & \multicolumn{1}{c}{RFE} \\
 \midrule
\multicolumn{10}{c}{\textbf{F1 class 0}}\\
\midrule
c0 & 0.65 & 0.74 & 0.69 & \textbf{0.75} & 0.67 & \textbf{0.75} & 0.74 & 0.56 & 0.66 \\
c1 & \textbf{0.99}\textsuperscript{a} & \textbf{0.99}\textsuperscript{a} & \textbf{0.99}\textsuperscript{a} & \textbf{0.99}\textsuperscript{a} & 0.43 & \textbf{0.99}\textsuperscript{a} & -\textsuperscript{b} & 0.01 & \textbf{0.99}\textsuperscript{a} \\
c2 & \textbf{0.97} & \textbf{0.97} & 0.93 & 0.96 & 0.85 & 0.94 & 0.64 & 0.85 & \textbf{0.97} \\
c3 & \textbf{0.97} & \textbf{0.97} & -\textsuperscript{b} & 0.93 & 0.89 & 0.93 & 0.95 & 0.87 & 0.90 \\
c4 & \textbf{0.72} & \textbf{0.72} & 0.71 & \textbf{0.72} & 0.01 & \textbf{0.72} & -\textsuperscript{b} & 0.27 & 0.25 \\
c5 & \textbf{0.99} & \textbf{0.99} & 0.98 & \textbf{0.99} & 0.96 & \textbf{0.99} & \textbf{0.99} & \textbf{0.99} & 0.98 \\
\midrule
\multicolumn{10}{c}{\textbf{F1 class 1}}\\
\midrule
c0 & 0.63 & \textbf{0.75} & 0.72 & \textbf{0.75} & 0.72 & \textbf{0.75} & \textbf{0.75} & 0.53 & 0.7 \\
c1 & \textbf{0.99}\textsuperscript{a} & \textbf{0.99}\textsuperscript{a} & \textbf{0.99}\textsuperscript{a} & \textbf{0.99}\textsuperscript{a} & 0.76 & \textbf{0.99}\textsuperscript{a} & -\textsuperscript{b} & 0.70 & \textbf{0.99}\textsuperscript{a} \\
c2 & \textbf{0.97} & \textbf{0.97} & 0.93 & 0.96 & 0.87 & 0.95 & 0.79 & 0.85 & \textbf{0.97} \\
c3 & 0.98 & \textbf{0.99} & -\textsuperscript{b} & 0.96 & 0.94 & 0.96 & 0.97 & 0.93 & 0.95 \\
c4 & 0.50 & 0.47 & 0.51 & 0.47 & 0.67 & 0.47 &-\textsuperscript{b} & \textbf{0.69} & \textbf{0.69} \\
c5 & \textbf{0.94} & 0.92 & 0.71 & 0.90 & 0.26 & 0.85 & 0.93 & 0.86 & 0.74 \\
\midrule
\multicolumn{10}{c}{\textbf{MCC}}\\
\midrule
c0 & 0.29 & \textbf{0.50} & 0.41 & \textbf{0.50} & 0.38 & \textbf{0.50} & \textbf{0.50} & 0.10 & 0.36 \\
c1 & \textbf{0.99}\textsuperscript{a} & \textbf{0.99}\textsuperscript{a} & \textbf{0.99}\textsuperscript{a} & \textbf{0.99}\textsuperscript{a} & 0.39 & \textbf{0.99}\textsuperscript{a} & -\textsuperscript{b} & 0.04 & \textbf{0.99}\textsuperscript{a} \\
c2 & 0.94 & \textbf{0.95} & 0.86 & 0.92 & 0.72 & 0.89 & 0.54 & 0.70 & 0.94 \\
c3 & 0.95 & \textbf{0.96} & -\textsuperscript{b} & 0.89 & 0.83 & 0.89 & 0.92 & 0.80 & 0.86 \\
c4 & \textbf{0.33} & \textbf{0.33} & \textbf{0.33} & \textbf{0.33} & 0.00 & \textbf{0.33} & -\textsuperscript{b} & 0.23 & 0.23 \\
c5 & \textbf{0.93} & 0.91 & 0.70 & 0.89 & 0.28 & 0.84 & 0.92 & 0.86 & 0.71
\end{tabular}%

\label{tab:resultsClass}
\end{table*}

\begin{table*}[ht]
\caption{RMSEP and $R^2$ results for regression datasets.}
\centering
\begin{tabular}{cclcllll}
 & $M^\circ$ & \multicolumn{1}{c}{RENT} & StabSel & \multicolumn{1}{c}{RFR} & \multicolumn{1}{c}{L-score} & \multicolumn{1}{c}{mRMR} & \multicolumn{1}{c}{RFE} \\
 \midrule
\multicolumn{8}{c}{\textbf{RMSEP}}\\
\midrule
r0 & \multicolumn{1}{l}{\textbf{21.73}} & \multicolumn{1}{r}{24.01} & 79.21 & 150.98 & 237.37 & 93.11 & 123.22 \\
r1 & \multicolumn{1}{l}{\textbf{0.12}} & 0.15 & 0.16 & 0.16 & 0.32 & 0.17 & 0.23 \\
\midrule
\multicolumn{8}{c}{$\bm{R^2}$}\\
\midrule
r0 & \textbf{0.99} & \textbf{0.99} & 0.86 & 0.47 & -0.30 & 0.80 & 0.65 \\
r1 & \textbf{0.82} & 0.71 & 0.68 & 0.66 & -0.35 & 0.62 & 0.34
\end{tabular}%
\label{tab:resultsReg}
\end{table*}

For classification problems, the results achieved with RENT feature selection are competitive with the best results of the other methods, yielding better or equally high F1 scores for predicting class 0 in five out of six datasets. For c0, the performance is only $0.01\%$ below the top value of $0.75$. Also, for class 1, RENT achieves the highest performance for four out of six datasets. For dataset c5 the performance is close to the best F1 score. MCC, which is more robust than the F1 score for unbalanced class settings, is highest for five out of six datasets with RENT feature selection. For c5, $M^\circ$ has a higher MCC, but with a much higher number of features (593 features) than RENT (16 features). With the regression datasets, RENT achieves a performance superior to StabSel, RFR, L-score, mRMR, and RFE and competitive performance to $M^\circ$ for both measures, RMSEP and $R^2$. 

In many cases, $M^\circ$ is not able to restrict the number of features as efficiently as RENT. Using the same regularization parameters as for RENT, $M^\circ$ selects the following number of features: 290 (c1) vs. 37 for RENT and 593 (c5) vs. 16 for RENT, respectively (see Table 2).

Overall, StabSel achieves good results for all datasets underlining the merits of ensemble feature selection concepts. Note that no results could be obtained for dataset c3 since no feature reached a sufficient selection frequency across all models.
The random forest yields competitive results for most datasets in classification (RFC) setups but performs noticeably worse in regression (RFR) setups. 
L-score and mRMR appear to provide weak performance scores compared to their competing feature selectors. For L-score, the low scores can be explained by its unsupervised setup, which makes it harder to relate the model to any target variable. With mRMR, especially the performances for c1, c2, and c4 are weak. For c1 and c4 this weakness can partly be attributed to the available implementation, which produced an error for these datasets (denoted with superscript \textsuperscript{b} in Table \ref{tab:resultsClass}).
On the other hand, F-score performs well, especially for predicting class 0. The reliefF method achieves good results for c5 but is among the poorest feature selectors for c0, c1, c2 and c3. Neither F-score nor reliefF are applicable to regression problems using the available implementations. 
Dataset c4 is of particular interest since opposite behavior can be observed among the feature selectors. RENT, StabSel, RFC, and F-score perform well when predicting class 0, whereas the other methods achieve higher scores for class 1. Hence, we assume that the features selected from c4 introduce a bias towards class 0 or class 1, respectively. In terms of MCC, which accounts for both classes in parallel, the best results are achieved by RENT, StabSel, RFC, and F-score.

\begin{figure}[ht]
    \centering
    \includegraphics[width=0.7\columnwidth, clip]{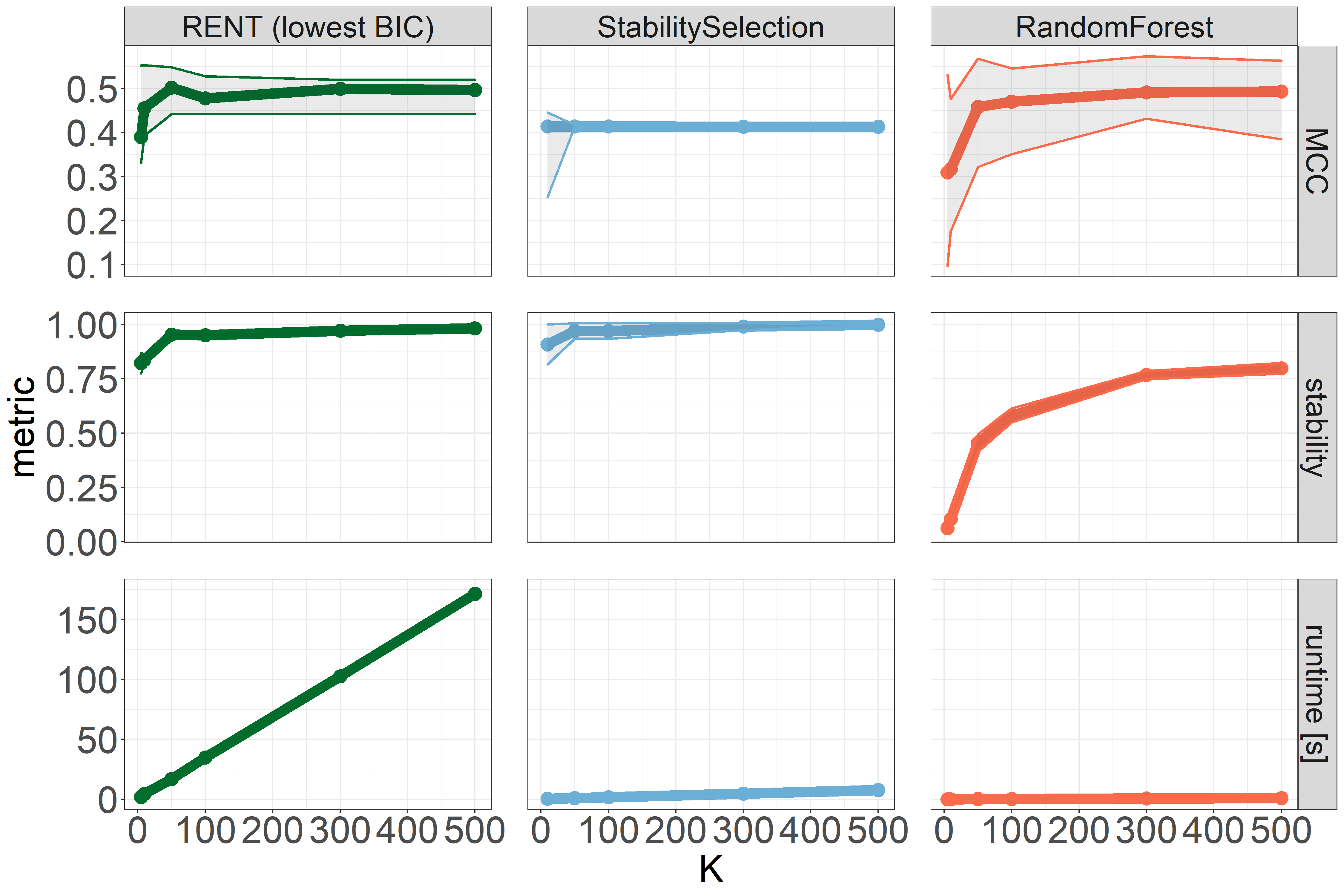}%
    \caption{Comparison of stability, performance, and runtime of three ensemble based feature selectors, where $K$ is the number of elementary models in RENT.}
    \label{fig:stability}
\end{figure}

Fig. \ref{fig:stability} depicts the experimental results for comparing the ensemble feature selectors with varying $K$, given the same setup as Fig. \ref{fig:stability_RENT}. 
While RFC achieves a similar performance as RENT, it is the most unstable ensemble approach for lower number of trees $K$. Even for high $K$, RFC never achieves the same stability as RENT and StabSel. Furthermore, RFC has the highest variance in MCC. On the other hand, StabSel reaches higher stability but lower MCC scores than its competitors between $K=50$ and $K=100$. 
The provided stability analysis underlines the strong properties of RENT, compared to random forests, which are known to be unstable in multiple scenarios \cite{calle2011letter}.

Regarding the computational costs\footnote{All results were acquired by running R 4.1.1 and Python 3.8.10 on a Windows 10 machine with a 4-core Intel i5 CPU \@ 1.8 GHz and 512 GB RAM.} of the ensemble feature selectors in our study, Fig. \ref{fig:stability} demonstrates that the runtime increases linearly in $K$ for all methods. RENT takes longer to compute which might be caused by the fact that the implementation does not yet exploit the full potential for runtime optimization and different implementations and programming languages were used for elementary operations. 

\section{Exploratory Post-Hoc Analysis}
As an ensemble model approach, RENT offers additional information that can be integrated into exploratory post-hoc analyses. The two post-hoc analyses presented in this section give the user tools to 1) further investigate objects in the dataset and identify which of those are difficult to predict and which not; 2) exploit this information in a principal component analysis model trained on the selected features to understand why some objects are difficult to predict. 

\subsection{Analysis of Training Objects}
\label{subsec:obj_predictions}
Based on the ensemble of elementary models in RENT, it is possible to compute summary statistics on a single-object level. Such information may contribute to improved interpretability of the model in general and single objects in the data in particular. For this purpose, we analyze the predictions of individual objects across all models $M_k$, $k=1,\dots,K$. For binary classification problems, we observe the distribution of correct and incorrect classifications of single objects in $X_{val}^{k}$, and thereby gain insights into the consistency of assigning an object to its true class. From a statistical perspective, this means that we can identify objects with deviating properties belonging to the same class based on the information whether the label of an object is difficult to predict or not. For regression problems, we similarly use the mean absolute errors. Below, we will exemplify the proposed post-hoc analysis for dataset c3.

\begin{figure*}[ht]
    \centering
    \begin{subfigure}{0.45\columnwidth}
        \includegraphics[width=\textwidth, trim=50 0 100 77, clip]{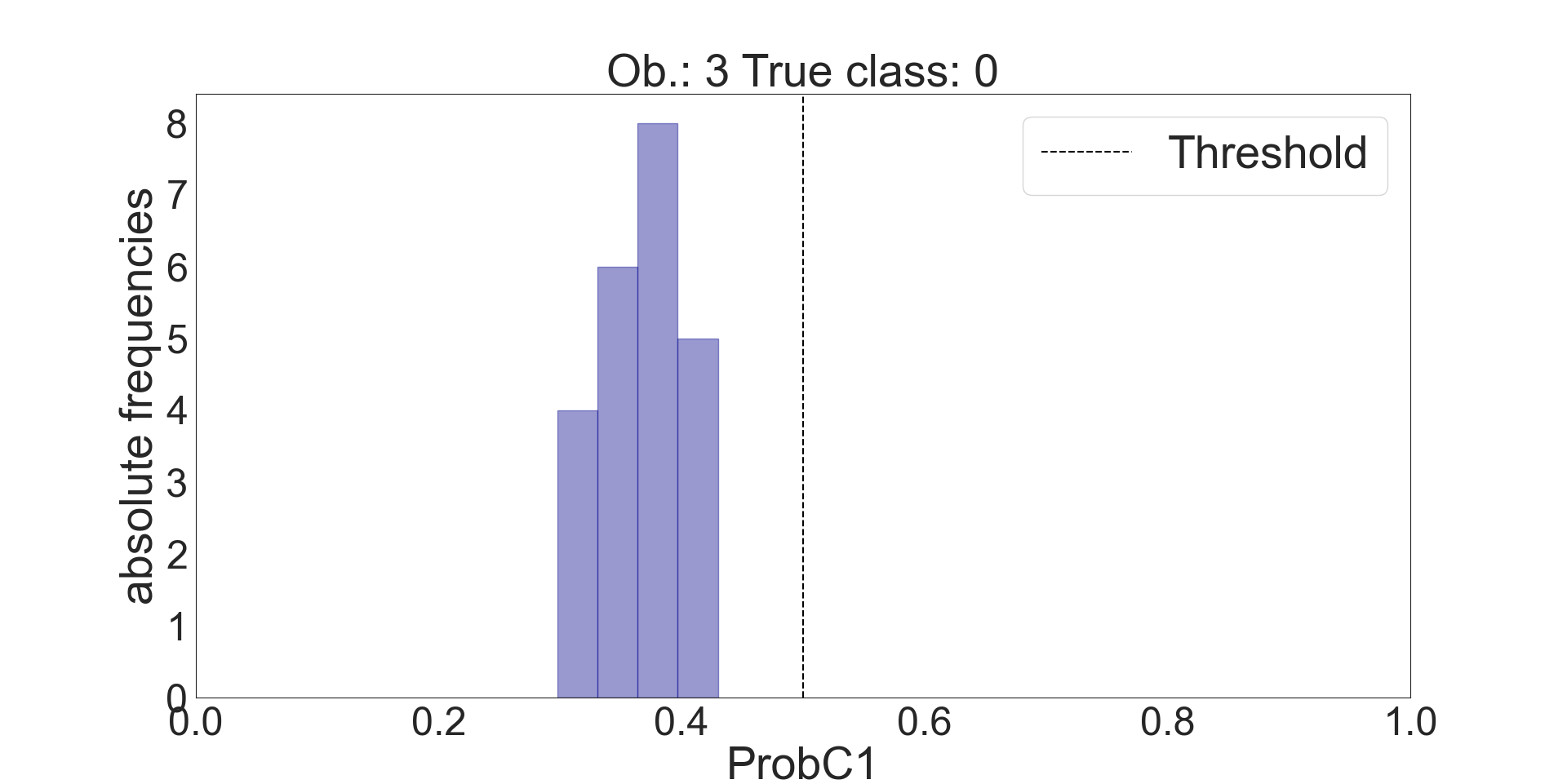}%
        \caption{Object no. 3, true class: 0.}
    \end{subfigure}
    \hfill
    \begin{subfigure}{0.45\columnwidth}
        \includegraphics[width=\textwidth, trim=50 0 100 77, clip]{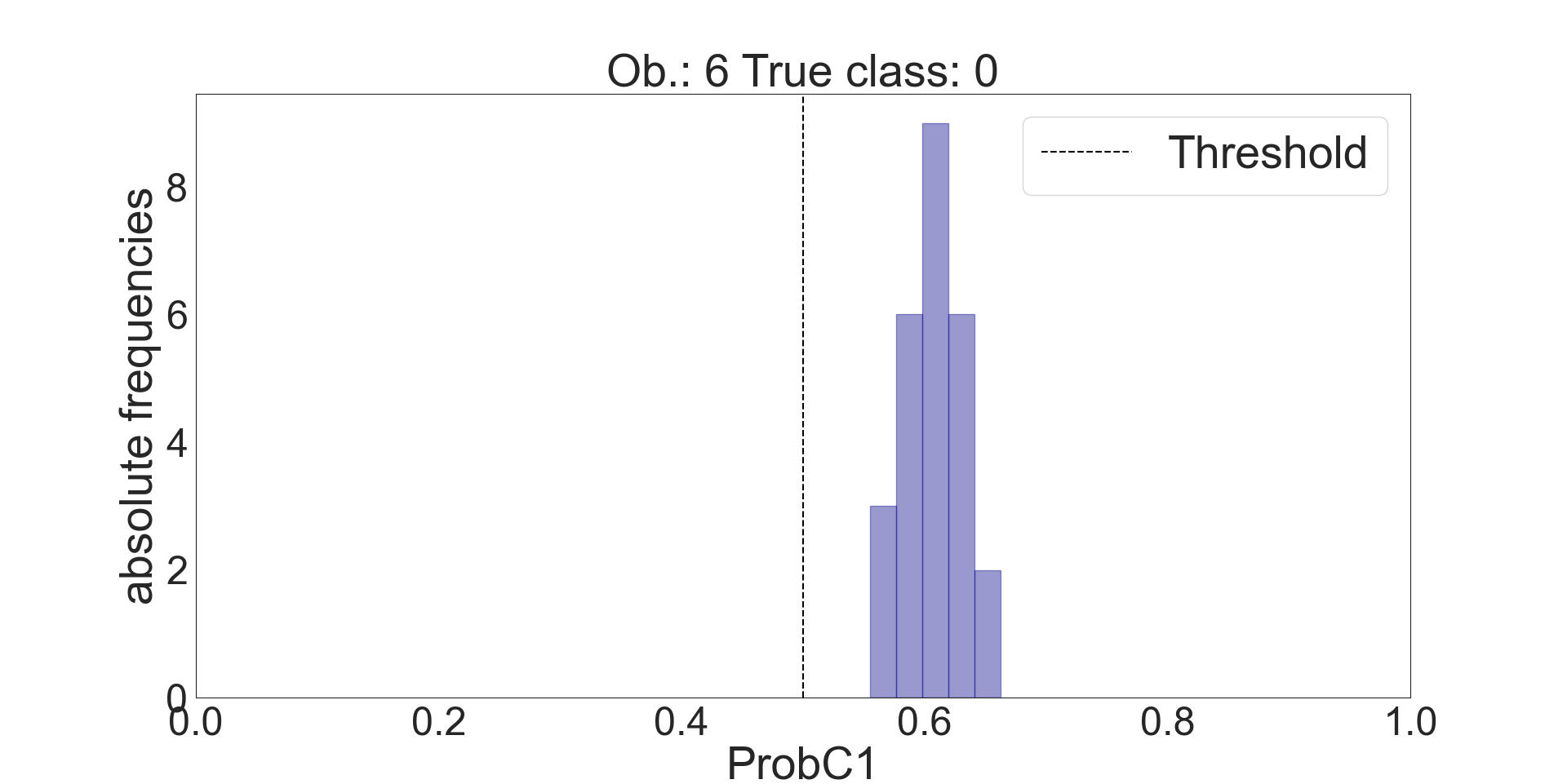}%
        \caption{Object no. 6, true class: 0.}
    \end{subfigure}\\
    \vspace{10pt}
    \begin{subfigure}{0.45\columnwidth}
        \includegraphics[width=\textwidth, trim=50 0 100 77, clip]{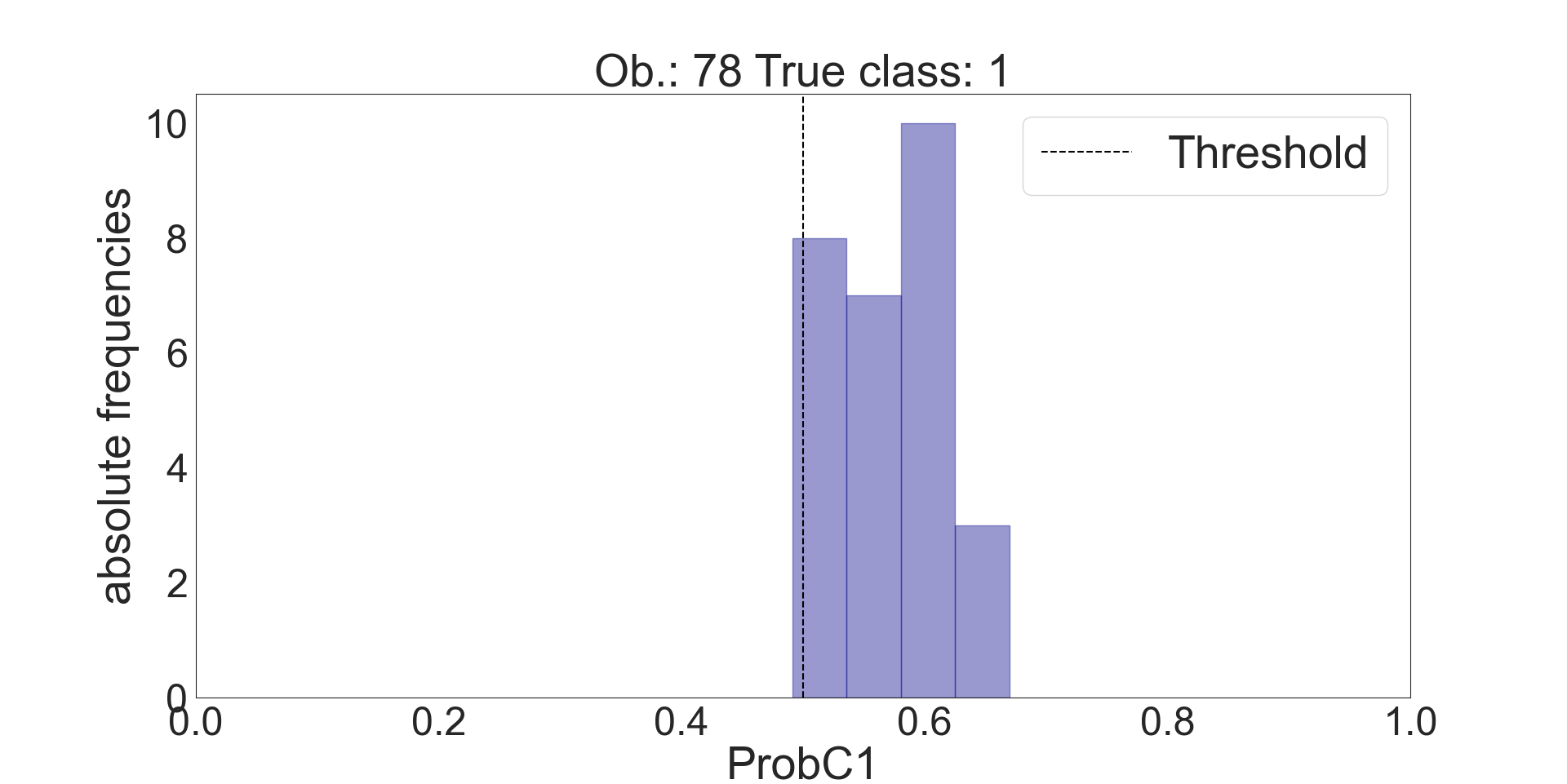}%
        \caption{Object no. 78, true class: 1.}
    \end{subfigure}    
    \hfill
    \begin{subfigure}{0.45\columnwidth}
        \includegraphics[width=\textwidth, trim=50 0 100 77, clip]{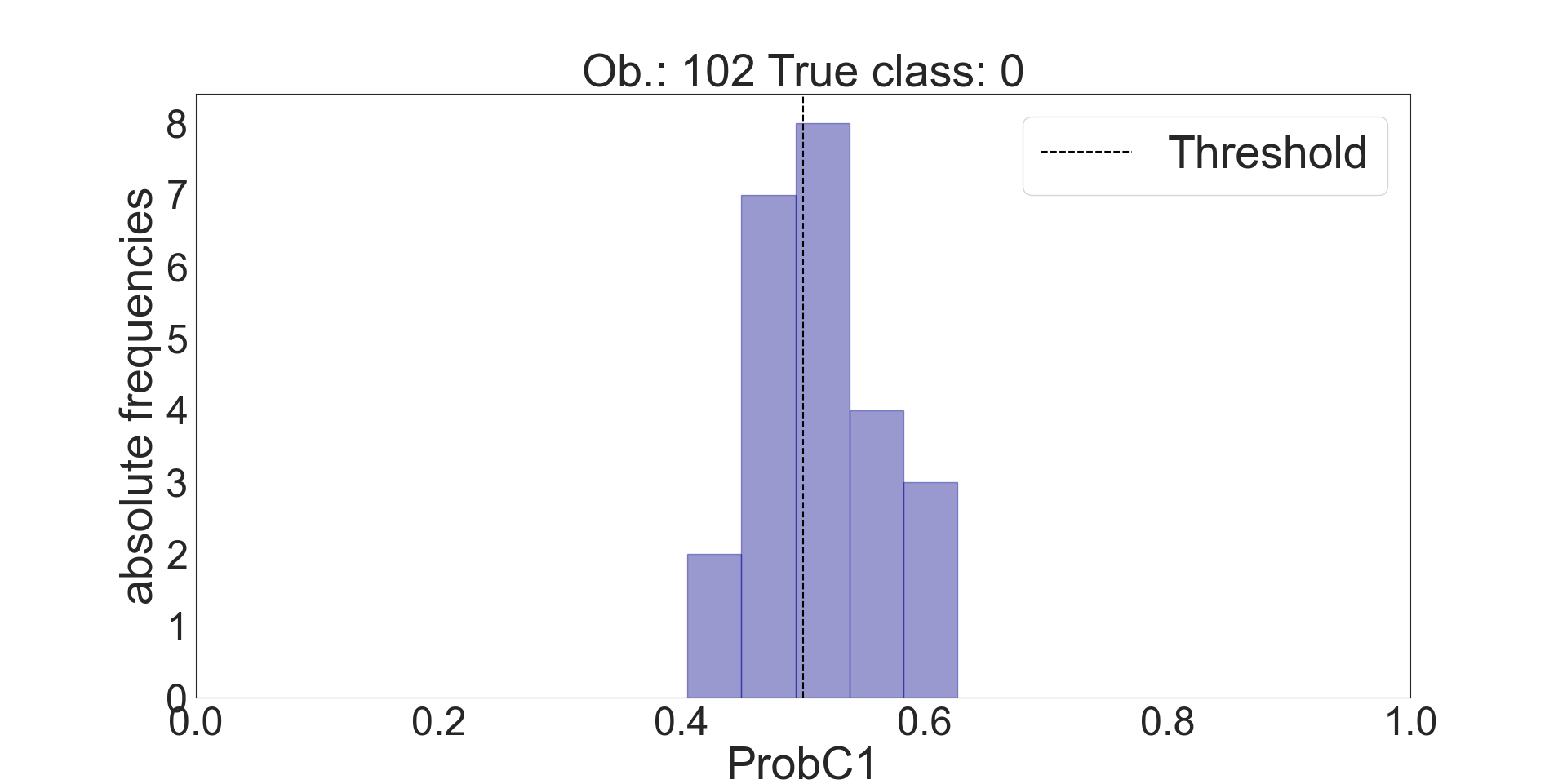}%
        \caption{Object no. 102, true class: 0.}
    \end{subfigure}
    \caption{Distribution of the class probability $\hat{y}_i$ of $\bm{x}_i$ being assigned class 1 (ProbC1s) for objects (patients) 3, 6, 78 and 102 estimated by $K$ different models of RENT. The first axis shows the ProbC1s, the second axis shows the absolute frequencies.}
    \label{fig:incorr}
\end{figure*}

Given an object $\bm{x}_i\in X_{val}^{k}$, the logistic regression model $M_k$ outputs a class probability $\hat{y}_i$ of $\bm{x}_i$ being assigned class 1 (ProbC1).
Among the $K$ models built within RENT, we obtain a ProbC1 value each time an object $\bm{x}_i$ appears in $X_{val}^{k}$, $k=1,\dots,K$. Aggregating this information by object, we can derive statistics and describe the distribution of the ProbC1s for each object $\bm{x}_i\in X_{val}^{k}$ by a histogram, as shown in Fig. \ref{fig:incorr}. 
These results are generated from dataset c3 (Breast cancer Wisconsin), where we denote a single object in the dataset as a cancer patient. Incorrect predictions provide evidence for patients that are hard to classify or show different characteristics compared to patients from the same class that are easy to classify. We observe that patient 3 belongs to class 0 and that the predicted probabilities of patient 3 are consistently below 0.5, which is the standard decision boundary for logistic regression models. In other words, patient 3 is predicted correctly every time she is part of $X_{val}^{k}$. Patient 6 belongs to class 0; however, the predicted probabilities are consistently above 0.5, meaning that her class label is always mispredicted. For patients 78 and 102 we observe probabilities both above and below 0.5, indicating that the class predictions of these two patients are rather uncertain, however, to a different degree. Fig. \ref{fig:incorr} reflects the detailed information provided in Table \ref{tab:incorrpred}. 

\begin{table}[ht]
\caption{In-depth analysis of predictions for four patients from the Breast cancer Wisconsin dataset (dataset c3), see Fig. \ref{fig:incorr}. $\#\;val\;set$ denotes how often the object was part of a validation set (between 1 and $K=100$), $true\;class$ is the true class, $\#\;incorrect$ describes how often the object was incorrectly predicted and $\%\; incorrect$ is the corresponding percentage.}
\centering
\begin{tabular}{c | c c c c }
        \toprule
        object & \# val set & true class & \# incorrect & \% incorrect\\
         \midrule
         3 & 23 & 0 & 0 & 0\\
         6 & 26 & 0 & 26 & 100\\
         78 & 28 & 1 & 2 & 7.1 \\
         102 & 24 & 0 & 13 & 54.2\\
        \bottomrule
\end{tabular}
\label{tab:incorrpred}
\end{table}
With a \textit{\% incorrect} of $54.2\%$, the class predictions for patient 102 are extremely unstable among the 24 models, where this patient was part of the validation set. This type of information on prediction stability may provide a good starting point for detailed studies on how a hard-to-classify object differs from objects that are consistently assigned to the correct class. Thus, it could be of high relevance, inter alia for medical experts, who may identify patients with deviating data characteristics. These difficulties may arise from dominating phenomena in the measured features or measurement errors.

In this way, ensemble based approaches such as RENT allow in-depth analysis of the distribution of class probabilities rather than restricting to single class predictions. 

\subsection{Principal Component Analysis (PCA) on Selected Features}
\begin{figure*}[ht]
 \centering
 \begin{subfigure}{0.45\columnwidth}
    \centering
    \includegraphics[height = 140px]{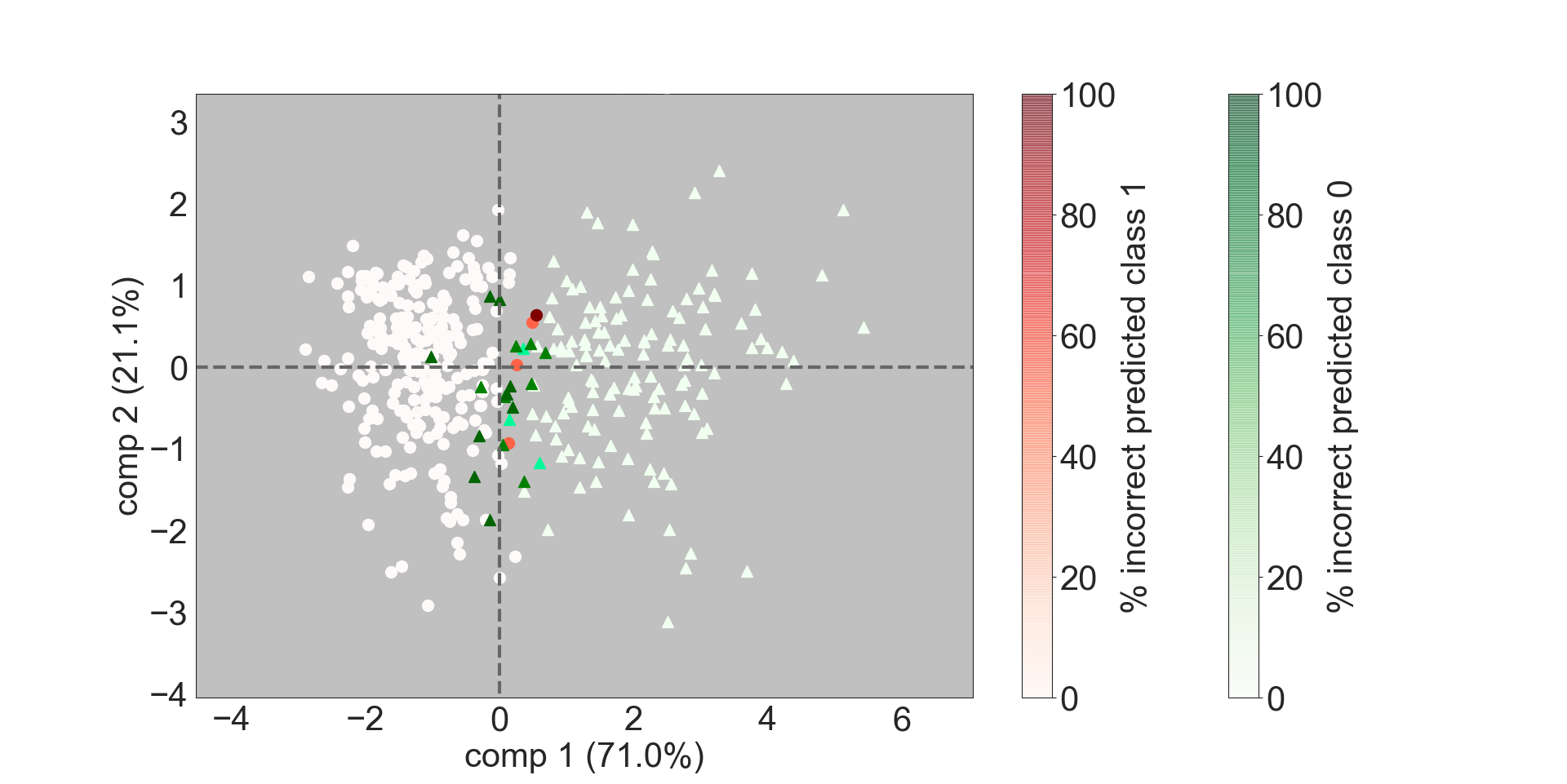}
    \caption{The shape of the PCA scores identifies the true class (class 0: triangles; class 1: circles).}
    \label{fig:PCA}
    \end{subfigure} \hfill
    \begin{subfigure}{0.45\columnwidth}
    \centering
   \includegraphics[height = 140px, trim=30 0 180 0, clip]{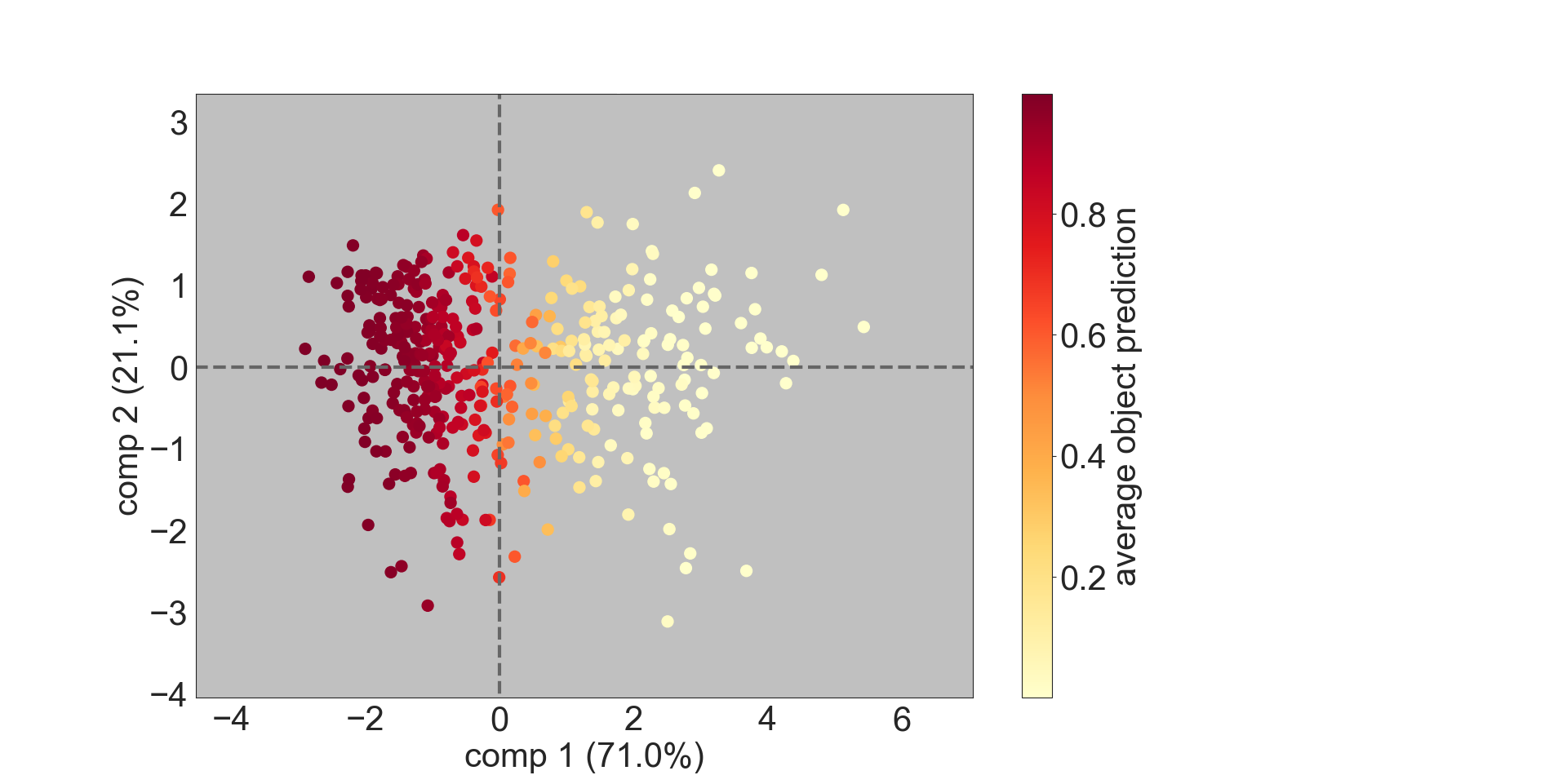}
   \caption{PCA scores colored by the relative frequency that an object belongs to class 1 across all elementary models. }
    \label{fig:PCA_scores_pred}
 \end{subfigure}
 \begin{subfigure}{0.45\columnwidth}
    \centering
    \includegraphics[height = 140px, trim=10 0 400 0, clip]{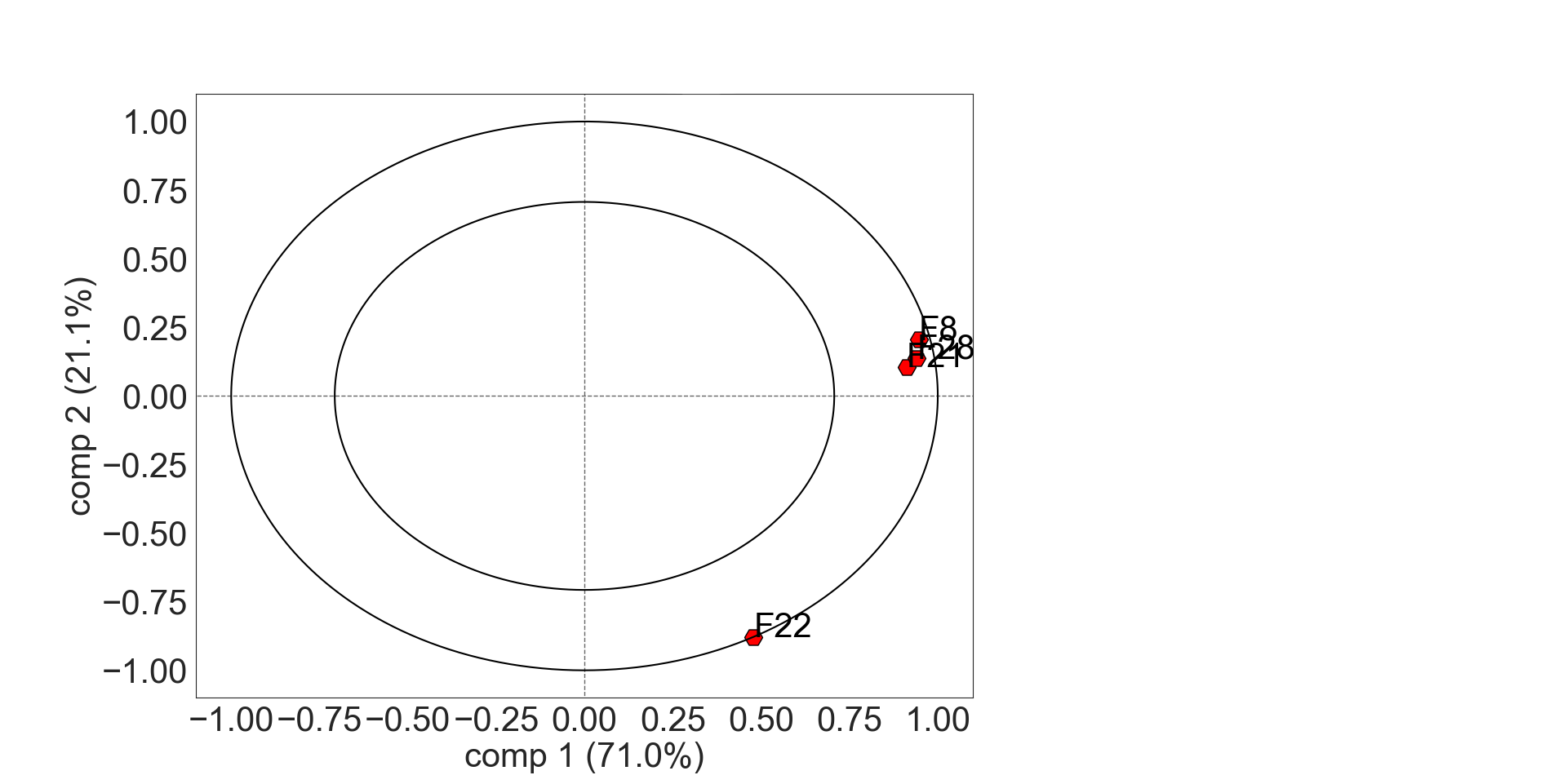}
    \caption{Correlation loadings plot.}
    \label{fig:PCA_loadings}
 \end{subfigure}
 \caption{PCA scores and correlation loadings of the Breast cancer Wisconsin dataset after RENT feature selection. The scores in Fig. \ref{fig:PCA} and \ref{fig:PCA_scores_pred} provide an overview of how the objects are distributed in the subspace spanned by components 1 and 2. The correlation loadings in Fig. \ref{fig:PCA_loadings} indicate how the selected features contribute to the variance explained by components 1 and 2.}
\end{figure*}

By a PCA \cite{PCA_mardia} of $X_{train}^{\star}$, we can obtain a better understanding of the properties of objects and their relation to the features selected by RENT. Note, that unlike in machine learning, where PCA is often only used for feature extraction, visualization (plotting) of PCA scores, PCA loadings, and PCA correlation loadings \cite{martens_correlation_loadings} may be efficient for the purpose of model interpretation. Fig. \ref{fig:PCA} and Fig. \ref{fig:PCA_scores_pred} show the scores \footnote{The calculations rely on the PCA implementation provided in the Python package "hoggorm" package \cite{hoggorm}.} of the first two principal components (comp 1 and comp 2) applied to the Breast cancer Wisconsin dataset, but with hues based on different information acquired from the ensemble. Every data point in the scores plot represents one object in the data or specifically to this dataset, one patient. The first two principal components explain 92.1\% of the total variance in the data that are contained in the selected features. The remaining $7.9\%$ are explained by the remaining principal components. In particular, Fig. \ref{fig:PCA} shows how the objects are distributed in the sub-space spanned by components 1 and 2. 
The two classes are well separated, with the objects of class 1 (circles) on the left side of the plot and class 0 (triangles) on the right side. Using results from RENT, the information in this plot may be further enhanced by coloring each object by its true class (class 0 - green triangles; class 1 - red circles) graded according to \textit{$\%$ incorrect} in Table \ref{tab:incorrpred}. Higher color saturation refers to a higher percentage of incorrect label predictions---suggesting that the object shows an anomalous behavior, which the model cannot sufficiently cover. In this example, objects with ambiguous classes accumulate in the middle area, most of them are close to the intuitive decision boundary. Fig. \ref{fig:PCA_scores_pred} shows the same scores as seen in Fig. \ref{fig:PCA}. However, the objects are colored by their average ProbC1 (the average probability of an object belonging to class 1). Again, objects with either a very high or a low value cluster on the right and the left-hand side, respectively. Objects---which we know are difficult to classify---with scores for comp 1 ranging between $-0.5$ and $1$ are located in the center of the image. PCA can also be performed on each class separately, to investigate within-class variations, as shown in extensive Jupyter notebook examples in the RENT GitHub repository.

In addition to the PCA scores, the correlation loadings plot is shown in Fig. \ref{fig:PCA_loadings}, where every point represents one feature in the plane spanned by comp 1 and comp 2. The correlation loadings plot encodes 1) the level of contribution of the selected features to each of the components 1 and 2, and 2) how much of the variance in each feature is explained by the two components. The further away a correlation loading is located from the origin, the higher the amount of explained variance for the feature it represents. The inner and outer circles represent $50\%$ and $100\%$ of explained variances, respectively. Among the four selected features, feature 8, 21 and 28 contribute most to the first component that separates the two classes. It is also evident that these three features are highly correlated, as they are located so close to each other in Fig. \ref{fig:PCA_loadings}. Moreover, they are close to the outer ring, meaning that comp 1 explains nearly $100\%$ of the variance in those features. Feature 22 contributes to both components 1 and 2, but is the feature that contributes most to component 2. By superimposing the scores onto the correlation loadings plots, we can gather information on how the scores and features are interrelated. Features 8, 21 and 28 and objects of class 0 are in the same regions (right side)---indicating that objects of class 0 have high values for these features, while objects of class 1 on the opposite side (left side) have low values for those features

The above examples of post-hoc analysis illustrate how combining ensemble information with exploratory analysis by PCA, can provide deeper insight into the data.

\section{Discussion}\label{sec:Discussion}

In summary, RENT performs well on all experimental datasets presented in this study when compared to the other feature selection methods. 
In particular, a good trade-off between predictive performance and selection stability is achieved. We observe that 1) RENT is consistently among the best performing methods, 2) if outperformed by others, the difference in performance is mostly negligible, and 3) the often lower number of features selected by RENT is a clear benefit. RENT does not fail for any of the presented datasets, whereas other methods show weaknesses on at least one dataset, with regard to either a very large number of selected features or poor predictive quality. In particular, RENT consistently performs well, whether the data are long-thin---many objects compared to the number of features---or short-wide---relatively few objects compared to the number of features. The initial intention of RENT was to target short-wide datasets, which are particularly challenging when it comes to feature selection. In the presented evaluations, datasets c4 (Dexter text classification), c5 (OVA Lung), and r1 (Milk proteins) clearly fall into this category. 

In addition to competitive performance, the number of features selected by RENT for the studied datasets is comparably low, which is a strength of RENT in terms of interpretability of results. Furthermore, the object-wise visualization demonstrated in Section \ref{subsec:obj_predictions} can provide previously unseen insights into the properties of the dataset, which may be particularly relevant for medical applications, but also for many other applications in general.

Robustness with regard to noisy data is another strength of RENT, which can be achieved by the extensive use of drawing subsamples from the training set. Particularly the baseline model $M^{\circ}$, which is used as a benchmark in the experiments and achieves high performance on multiple datasets, is susceptible to poor initializations and hence, potentially less reliable for the selection of features. Although computationally more intensive than the comparing methods, RENT is less susceptible to poor initializations or convergence issues of optimization routines compared to other approaches. 

In total, RENT has five model parameters to adjust by the user: two account for regularization intensity ($\gamma$ and $\alpha$) and three cutoffs control the strictness of feature selection ($t_1$, $t_2$ and $t_3$). Both sets of hyperparameters are related, since a softer regularization allows a larger number of features, requiring higher cutoffs (and vice versa). Based on the presented parameter selection procedure using BIC, feature selectors which deliver a low number of features are favored in both stages. 

In the current formulation, RENT is applicable for binary classification and regression problems. As introduced in \cite{Gao}, multiclass feature selection is not trivial and will be part of further research. However, a multiclass classification problem can be split into several binary problems, using schemes such as one-vs-one (OVO), one-vs-all (OVA), or error-correcting output coding (ECOC), as described in \cite{schrunner18}.

\section{Conclusion}\label{sec:Conclusion}
In this work, we presented a feature selection technique for binary classification and regression problems. The algorithm builds on the idea of training multiple elastic net regularized models on unique training data subsets. In particular, we define feature importance criteria based on the empirical distribution of feature-wise model weights. Features are selected if their associated weights are regularly assigned high non-zero values with stable signs across the individual models of the ensemble. 

We provided experiments on datasets from different disciplines, demonstrating that RENT is effective with respect to quantitative performance measures and interpretability and robustness. For the presented setups, the stability is very high even with a moderate number of ensemble models and in five out of six binary classification datasets, RENT achieves the highest MCC scores compared to the established feature selectors used in this study. For the regression datasets, RENT performed better or almost equal to the competing approaches. Further, we showed how to utilize information from the ensemble of models in a post-hoc analysis, advancing single-object interpretability.

\section*{Acknowledgment}
This work was supported by the Norwegian Cancer Society [grant no. 182672-2016]. In special we thank Tormod Næs (Nofima) and Yngve Mardal Moe (University of Oslo) for their constructive discussions and valuable input to this work. Further, we thank the reviewers for their feedback which helped us to improve the paper. 

\bibliographystyle{unsrt}  

\bibliography{preprint}

\end{document}